\documentclass{article}
\pdfoutput=1

\usepackage[final]{neurips_data_2022}
\usepackage[utf8]{inputenc} % allow utf-8 input
\usepackage[T1]{fontenc}    % use 8-bit T1 fonts
\usepackage[colorlinks]{hyperref}       % hyperlinks
\usepackage{url}            % simple URL typesetting
\usepackage{booktabs}       % professional-quality tables
\usepackage{amsfonts}       % blackboard math symbols
\usepackage{nicefrac}       % compact symbols for 1/2, etc.
\usepackage{microtype}      % microtypography
\usepackage{xcolor}         % colors
\usepackage{comment}
\usepackage{graphicx}
\usepackage{subfig}
\usepackage{pgfplots} 
\usepackage{titletoc}
\usepackage{listings}
\usepackage{algorithm}% http://ctan.org/pkg/algorithm
\usepackage{algpseudocode}% http://ctan.org/pkg/algorithmicx
\pgfplotsset{compat=1.16}
\usepackage{footmisc}

\usepackage{ragged2e,array,longtable}
\newcolumntype{P}[1]{>{\RaggedRight\arraybackslash}p{#1}}
\newcolumntype{T}{c<{\ttfamily}}
\title{The BigScience ROOTS Corpus: \\ A 1.6TB Composite Multilingual Dataset}

\lstdefinestyle{htmlcssjs} {%
  basicstyle={\footnotesize\ttfamily},   
  xleftmargin={0.75cm},
  identifierstyle=\color{black},
  keywordstyle=\color{blue}\bfseries,
  ndkeywordstyle=\color{editorGreen}\bfseries,
  stringstyle=\color{editorOcher}\ttfamily,
  commentstyle=\color{brown}\ttfamily,
  language=HTML,
  alsodigit={.:;},	
  tabsize=2,
  showtabs=false,
  showspaces=false,
  showstringspaces=false,
  extendedchars=true,
  breaklines=true,
  literate=%
  {Ö}{{\"O}}1
  {Ä}{{\"A}}1
  {Ü}{{\"U}}1
  {ß}{{\ss}}1
  {ü}{{\"u}}1
  {ä}{{\"a}}1
  {ö}{{\"o}}1
}
\begin{document}

\maketitle
\vspace{-2cm} %remove empty space caused by an empty author command
\newcommand{\hfemoji}{\includegraphics[scale=0.035]{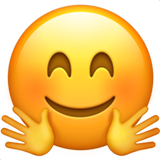}}
\newcommand{\bsemoji}{\includegraphics[scale=0.035]{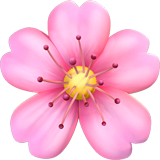}}

\begin{center}
% Added in first batch

% First authors: ordering TBD
\textbf{Hugo~Laurençon}$^{1*}$ \quad
\textbf{Lucile~Saulnier}$^{1*}$ \quad
\textbf{Thomas~Wang}$^{1*}$ \quad
\textbf{Christopher~Akiki}$^{2*}$ \quad
\textbf{Albert~Villanova~del~Moral}$^{1*}$ \quad
\textbf{Teven~Le~Scao}\textsuperscript{1*} \quad

% Co-authors
% Other Significant Code contributions: ordering TBD
\textbf{Leandro~von~Werra}$^{1}$ \quad
\textbf{Chenghao~Mou}$^{3}$ \quad % independent researcher
\textbf{Eduardo~González~Ponferrada}$^{4}$ \quad
\textbf{Huu~Nguyen}$^{5}$ \quad
\textbf{Jörg~Frohberg} $^{32}$ \quad
\textbf{Mario~Šaško} $^{1}$ \quad
\textbf{Quentin~Lhoest} $^{1}$ \quad

% Other co-authors: randomly ?
\textbf{Angelina~McMillan-Major}$^{1,6}$ \quad
\textbf{Gérard~Dupont}$^7$ \quad
\textbf{Stella~Biderman}$^{8,9}$ \quad
\textbf{Anna~Rogers}$^{10}$ \quad
\textbf{Loubna~Ben~allal}$^{1}$ \quad
\textbf{Francesco~De~Toni}$^{11}$ \quad
\textbf{Giada~Pistilli}$^{1}$ \quad
\textbf{Olivier~Nguyen} $^{28}$ \quad
\textbf{Somaieh~Nikpoor}$^{12}$ \quad
\textbf{Maraim~Masoud}$^{13}$ \quad
\textbf{Pierre~Colombo}$^{14}$ \quad
\textbf{Javier~de~la~Rosa}$^{15}$ \quad
% \textbf{Giada~Pistilli}$^1$ \quad duplicate
\textbf{Paulo~Villegas}$^{16}$ \quad
\textbf{Tristan~Thrush}$^{1}$ \quad
\textbf{Shayne~Longpre}$^{17}$ \quad
\textbf{Sebastian~Nagel}$^{19}$ \quad
\textbf{Leon~Weber} $^{20}$ \quad
\textbf{Manuel~Romero~Mu\~{n}oz} $^{21}$ \quad
\textbf{Jian~Zhu} $^{22}$ \quad
\textbf{Daniel~van~Strien} $^{23}$ \quad
\textbf{Zaid~Alyafeai} $^{24}$ \quad
\textbf{Khalid~Almubarak} $^{25}$ \quad
\textbf{Vu~Minh~Chien} $^{26}$ \quad
\textbf{Itziar~Gonzalez-Dios} $^{27}$ \quad
\textbf{Aitor~Soroa} $^{27}$ \quad
\textbf{Kyle~Lo} $^{29}$ \quad
% \textbf{Jonathan Chang} $^$ \quad
\textbf{Manan~Dey} $^{30}$ \quad
\textbf{Pedro~Ortiz~Suarez} $^{31}$ \quad
\textbf{Aaron~Gokaslan} $^{18}$ \quad
\textbf{Shamik~Bose}$^{3}$ \quad
\textbf{David~Ifeoluwa~Adelani}$^{33}$ \quad
\textbf{Long~Phan}$^{34}$ \quad
\textbf{Hieu~Tran}$^{34}$ \quad
\textbf{Ian~Yu}$^{35}$ \quad
\textbf{Suhas~Pai}$^{36}$ \quad
\textbf{Jenny~Chim}$^{37}$ \quad

% Advisors
\textbf{Violette~Lepercq}\textsuperscript{1} \quad
\textbf{Suzana~Ilić}\textsuperscript{1} \quad
\textbf{Margaret~Mitchell}\textsuperscript{1} \quad
\textbf{Sasha~Luccioni}\textsuperscript{1} \quad
\textbf{Yacine~Jernite}\textsuperscript{1} \quad

% Affiliation
$^{1}$Hugging~Face \quad
$^2$Leipzig~University \quad
$^3$Independent~Researcher \quad
$^4$Ferrum Health \quad
$^5$Ontocord.ai \quad
$^6$University~of~Washington \quad
$^7$Mavenoid \quad
$^8$EleutherAI \quad
$^9$Booz~Allen~Hamilton \quad
$^{10}$University~of~Copenhagen \quad
$^{11}$University~of~Western~Australia \quad
$^{12}$CAIDP \quad
$^{13}$Independent~Researcher \quad
$^{14}$CentraleSupélec \quad
$^{15}$National~Library~of~Norway \quad
$^{16}$Telefonica~I+D \quad
$^{17}$MIT \quad
$^{18}$Cornell~University \quad
$^{19}$Common~Crawl \quad
$^{20}$Humboldt-Universität~zu~Berlin~and~Max~Delbrück~Center~for~Molecular~Medicine \quad
$^{21}$Narrativa \quad
$^{22}$University~of~Michigan,~Ann~Arbor \quad
$^{23}$British~Library \quad
$^{24}$King~Fahd~University~of~Petroleum~and~Minerals \quad
$^{25}$Prince~Sattam~bin~Abdulaziz~University~(PSAU) \quad
$^{26}$DETOMO~Inc. \quad
$^{27}$HiTZ~Center,~University~of~the~Basque~Country~(UPV/EHU) \quad
$^{28}$ServiceNow \quad
$^{29}$Allen~Institute~for~AI \quad
$^{30}$SAP \quad
$^{31}$Mannheim~University \quad
$^{32}$Apergo.ai \quad
$^{33}$Saarland~University \quad
$^{34}$VietAI~Research \quad
$^{35}$Aggregate~Intellect \quad
$^{36}$Bedrock~AI \quad
$^{37}$Queen~Mary~University~of~London \quad

$^{*}$ Equal contributions
\end{center}
% TODO: @thomas add correspondence

\begin{abstract}
As language models grow ever larger, the need for large-scale high-quality text datasets has never been more pressing, especially in multilingual settings. The BigScience workshop, a 1-year international and multidisciplinary initiative, was formed with the goal of researching and training large language models as a values-driven undertaking, putting issues of ethics, harm, and governance in the foreground. This paper documents the data creation and curation efforts undertaken by BigScience to assemble the \underline{R}esponsible \underline{O}pen-science \underline{O}pen-collaboration \underline{T}ext \underline{S}ources (\textbf{ROOTS}) corpus, a 1.6TB dataset spanning 59 languages that was used to train the 176-billion-parameter \underline{B}igScience \underline{L}arge \underline{O}pen-science \underline{O}pen-access \underline{M}ultilingual (\textbf{BLOOM})\citep{bigscience_workshop_2022} language model. We further release a large initial subset of the corpus and analyses thereof, and hope to empower large-scale monolingual and multilingual modeling projects with both the data and the processing tools, as well as stimulate research around this large multilingual corpus.
\end{abstract}

\tableofcontents
\newpage

%\begin{abstract}
%As language models grow ever larger, the need for large-scale high-quality text datasets has never been more pressing. This urgency is often met in practice by an unreflected use of Web data as ``model fodder'', relegating issues of ethics, harm, and governance to secondary positions of post-facto crisis management and damage control. This ``train first, ask questions later'' approach to science clashes with the value-driven approach of the BigScience workshop, a 1-year international and multidisciplinary initiative, formed with the goal of researching and training large language models. This paper documents the data creation and curation efforts undertaken by BigScience to assemble a 1.6TB dataset spanning 59 languages that was used to train the 176-billion-parameter \underline{B}igScience \underline{L}arge \underline{O}pen-science \underline{O}pen-access \underline{M}ultilingual language model (BLOOM). We further release a large initial subset of the corpus and analyses thereof, and hope to stimulate further research into studying this large multilingual corpus.
%\end{abstract}

% Introduction
\section{Introduction}\label{sec:intro}
%Work on the outline at: %\url{https://docs.google.com/document/d/15SzkO9BBtI_byqbF0i6GGRNsJq-J3PaUIoP0fgNPtu4/}
%Introduce the project, the requirements of the corpus, the motivating needs, and issues with other approaches.
%Sasha, Chris, Yacine (1 page)

BigScience\footnote{\href{https://bigscience.huggingface.co/}{https://bigscience.huggingface.co/}} started in May 2021 as a one-year long open collaborative research initiative that gathered over a thousand participants around the world to study large language models~(LLM). One of the founding goals of BigScience
%% Thomas: Remove, doesn't bring anything
%---a goal that would end up leading to collaboration across different working groups---
was to train an open-access, massively multilingual LLM, comparable in scale to GPT-3 \citep{brown2020} yet trained on a better documented and more representative multilingual dataset. The overall BigScience workshop was designed as a collaborative \citep{guiding-principles-participatory-nlp,envisioning-communities} and value-driven \citep{Birhane2021Values} endeavor. 
%While there has been extensive discussion of the importance of datasets, most relevant to the corpus creation aspect of our project is \cite{Scheuerman2021} who outline how data efforts in computer vision tend to prioritize ``\textit{efficiency [over] care; universality [over] contextuality; impartiality [over] positionality}\ldots.'' 
Throughout the process of building this corpus we engaged in simultaneous investigation of ethical \citep{talat2022you}, sociopolitical \citep{mcmillanmajor2022catalogue}, and data governance issues \citep{jernite2022governance} with the explicit goal of doing good for and by the people whose data we collected.

\begin{figure}
\centering
\includegraphics[width=0.7\textwidth]{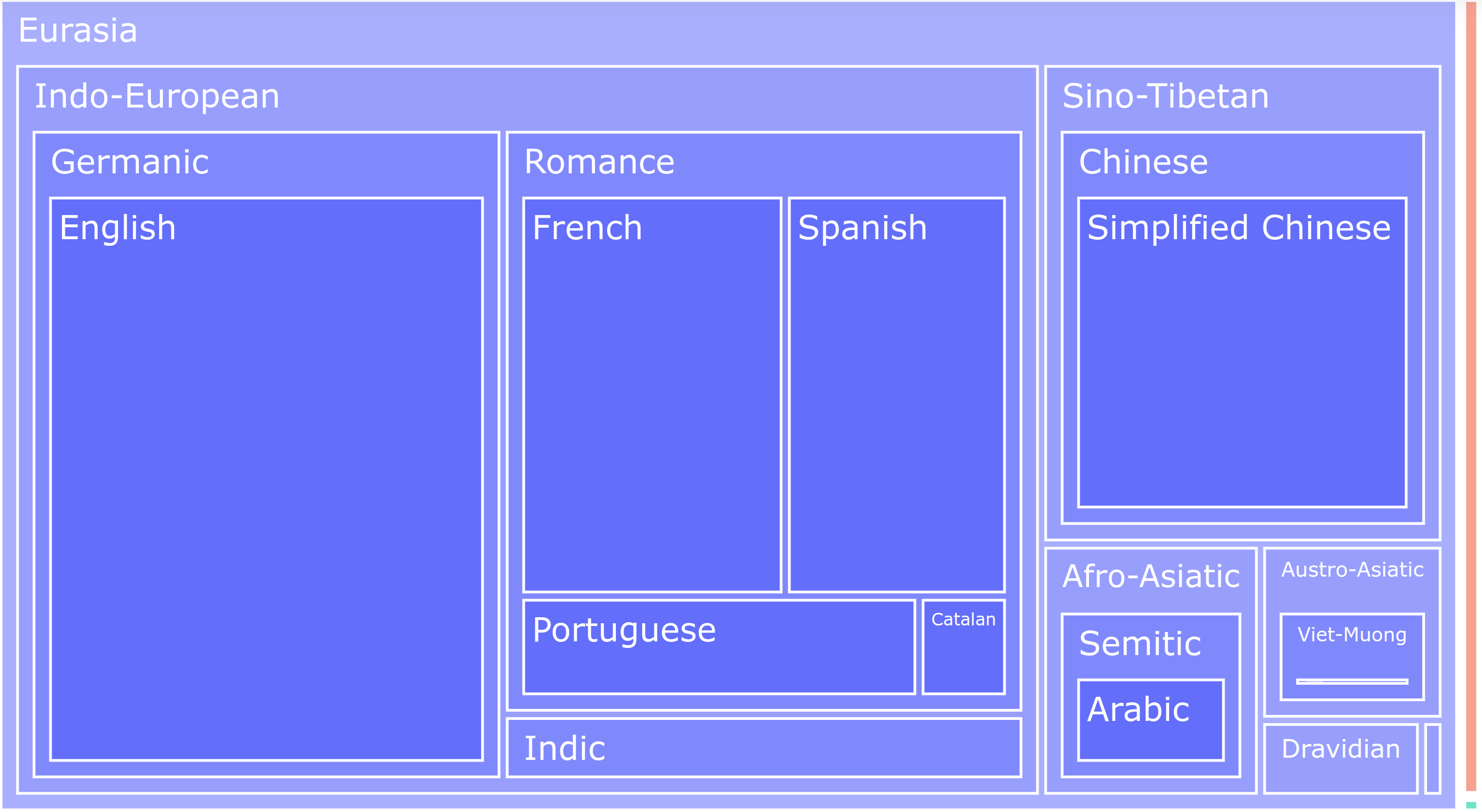}
\hfill
% \hspace{0.05\textwidth}
% \includegraphics[width=0.2\textwidth]{figures/code_waffle_vertical.png}
\includegraphics[width=0.21\textwidth]{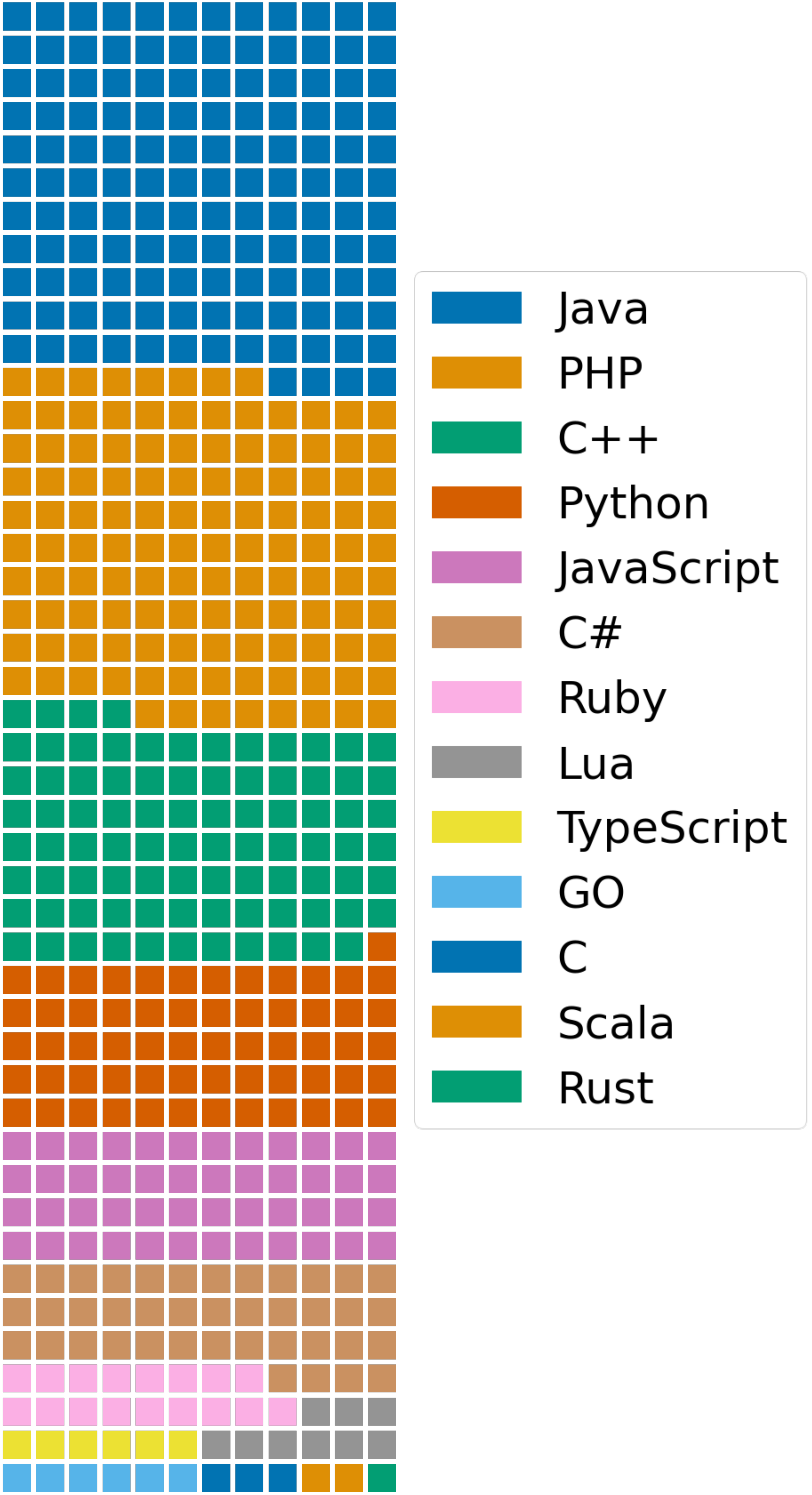}
\caption{
Overview of ROOTS.
Left: A treemap of natural language representation in number of bytes by language family. The bulk of the graph is overwhelmed by the 1321.89 GB allotted to Eurasia. The orange rectangle corresponds to the 18GB of Indonesian, the sole representative of the Papunesia macroarea, and the green rectangle to the 0.4GB of the Africa linguistic macroarea.
Right: A waffle plot of the distribution of programming languages by number of files. One square corresponds approximately to 30,000 files.
}
\label{fig:corpus_distribution}
\end{figure}

Sourcing and building the dataset was organized around four working groups: \textbf{Data Governance} which helped define the project's values and design our approach to data usage and release in an international context, \textbf{Data Sourcing and Preparation} which was tasked with overseeing data collection, curation efforts, and  \textbf{Privacy} for privacy risks and sanitizing the dataset, \textbf{Legal Scholarship} which helped define the multi-jurisdiction legal context in which the entire workshop was to operate, and we discuss practical implications throughout the paper where appropriate. An overview of the BigScience Corpus is provided in figure~\ref{fig:corpus_distribution}.

The goal of the current paper is twofold: (1)~we present a preliminary gated, subject to committing to the BigScience ethical charter\footnote{\label{charter}\href{https://hf.co/spaces/bigscience/ethical-charter}{https://hf.co/spaces/bigscience/ethical-charter}}, release of a large subset of ROOTS\footnote{\href{https://hf.co/bigscience-data}{https://hf.co/bigscience-data}} (2)~we release the numerous data tools\footnote{\href{https://github.com/bigscience-workshop/data-preparation}{https://github.com/bigscience-workshop/data-preparation}} that were developed along the way and enabled us to curate, source, clean and inspect all 498 constituent datasets that come together to constitute ROOTS. This includes a preliminary results of the analyses that are currently being developed to study the corpus.

% Background and Approach
\subsection{Outline of the Paper}
The remainder of this paper details our approach to curating a web-scale dataset covering 59 languages, 46 natural languages and 13 programming languages --- the language choice was chiefly driven by the communities who participated in the effort given the importance we placed on language expertise.
Our final corpus is made up of two main components: 62\% of the text comes from a community-selected and documented list of language data sources and its collection process is described in section~\ref{section:sourcing}, and 38\% consists of text extracted from a pre-processed web crawl, OSCAR (\cite{ortiz-suarez-etal-2020-monolingual}), filtered with the help of native speakers, which is described in section~\ref{section:common_crawl_section}.

\subsection{Related Work}

\paragraph{Large Language Models and Large Text Corpora} 

The current dominant paradigm in natural language processing relies heavily on pre-trained models: large language models that can then be fine-tuned on a downstream task \citep{Ulmfit, Bert} or even used as-is without additional data \citep{radford2019language, brown2020}. In this paradigm, performance is directly correlated on both the model size and the dataset size and quality \citep{scalinglaws}, with recent models trained on up to 1.4 trillion tokens \citep{chinchilla} and dataset creation pipelines representing a significant part of large language model projects. Most such datasets, however, are not released, hindering further research. Exceptions include the Pile \citep{Gao2020}, a curated corpus of datasets for language modeling that has become widely used for training state-of-the-art English-language models \citep{lieber2021jurassic,Smith2022,black2022gpt,zhang2022opt}, and C4 and mC4 \citep{raffel2020exploring, mt5}, which have powered the T5 family of models; CC100 \citep{unsupervised-cross-lingual-representation-learning-at-scale} which has seen heavy use for multilingual modeling; and OSCAR \citep{OSCAR}, which has enabled monolingual non-English models.

\paragraph{Tooling, Visualization, and Replication}

Upstream from the finalized training datasets is the issue of processing methods and pipelines: both the operations that the datasets go through and the engineering effort required to apply them at terabyte scales. Existing work tends to fall on a spectrum from no details at all \citep{brown2020} to detailed filtering instructions, with \citep{raffel2020exploring} or without the dataset release \citep{Gopher} to detailed filtering instructions with the accompanying code \citep{Gao2020, unsupervised-cross-lingual-representation-learning-at-scale, OSCAR}. Even when the code is released, it tends to be built and tailored for the project's purpose. Consequently, large projects that do not re-use an existing dataset outright usually build their own pipeline rather than re-use an existing one on new data. However, data tools that were built and packaged in order to be used for other projects exist, such as OSCAR's Ungoliant and Goclassy \citep{Ungoliant, OSCAR}, which provides a distributed Common Crawl processing pipeline; CCNet \citep{wenzek2020ccnet}, built for quality filtering of multilingual Common Crawl dumps; and OpenWebText \citep{Gokaslan2019OpenWeb}, enabling Reddit dump processing.

\paragraph{Documenting Textual Corpora in NLP} 

An inspiration for our work is a recent emphasis on a more in-depth documentation of what is included and what is not in the corpora used for training NLP models 
%% Thomas: Remove, saving space
% in general, and large language models in particular. 
. The most notable example of this is the Pile, for which the authors themselves analyze and document a variety of syntactic and semantic properties of the dataset including structural statistics (n-gram counts, language, document sizes), topical distributions across its components, social bias and sentiment co-occurrence, pejorative content, and information about licensing and authorial consent, in addition to releasing a datasheet \citep{Biderman2022}. Other LM pre-training datasets that have been documented and analyzed include C4 \citep{Dodge2021,luccioni2021s,Kreutzer2022}, OSCAR \citep{Kreutzer2022} and BookCorpus \citep{bandy2021addressing}
%% Thomas: Remove, saving space
%, although those analysis were conducted after release
. While this kind of documentation is far from standard practice, it is becoming increasingly common given recent calls for better documentation~\citep{rogers-2021-changing, bender2021} as well as empirical studies on data memorization in language models \citep{carlini2019secret,carlini2022quantifying}.

\section{(Crowd) Sourcing a Language Resource Catalogue}
\label{section:sourcing}

%leandro, Yacine, Angie, Francesco (3/4 to 1 page)
The first part of our corpus, accounting for 62\% of the final dataset size (in bytes), was made up of a collection of monolingual and multilingual language resources that were selected and documented collaboratively through various efforts of the BigScience Data Sourcing working group.
The first such effort consisted in creating a tool to support metadata collection through open submissions, called the \href{https://hf.co/spaces/bigscience/SourcingCatalog}{BigScience Catalogue} and running a series of hackathons in collaboration with locally-focused ML and NLP communities such as Masakhane, Machine Learning Tokyo and LatinX in AI where participants could add and document entries for their languages to the catalogue \citep{mcmillanmajor2022catalogue}. This yielded a set of 252 sources, including at least 21 per considered language category.
We focused on metadata collection as a way to support selection of the sources for the final dataset and documentation of the final dataset.
In parallel, working group participants gathered additional Arabic language resources in the Masader repository \citep{zaid2021}, and proposed a list of websites of interest to increase the geographical diversity of our English, Spanish, and Chinese language data.
Finally, in order to explicitly test large language models' ability to handle computer code along with natural language, 
%% Thomas: Replace, untrue
%we selected permissively licensed code data available on GitHub.
%% with
we selected code data available on GitHub and StackExchange.

\subsection{Obtaining Data from the Identified Resources}
\label{section:obtaining}

\paragraph{Gathering Identified Datasets and Collections.}

First, we leveraged the BigScience Catalogue %% Thomas: Remove as doesn't bring much
%(specifically resources tagged as ``Primary sources'' or ``Processed language datasets'') 
and the Masader repository to start obtaining text from identified sources, which included both existing NLP datasets and collections of documents of various compositions.
Given the diversity of sources, hosting methods, data custodians, and formats, collecting this text required a collaborative effort.
To that end, we established a 2-phase approach: first, collect as many data sources as possible in an easily accessible location; second, map all of them to a common format to ease further processing.
%% Thomas: Remove, saving space,
% and use in training a language model easier

In the first phase, we organized \href{https://github.com/bigscience-workshop/data_tooling/wiki/datasets-hackathon}{an open hackathon} to start gathering identified sources on the Hugging Face Datasets hub~\citep{datasets}, in a dedicated organization\footnote{\href{https://hf.co/bigscience-catalogue-data}{https://hf.co/bigscience-catalogue-data}} (in order to manage access controls).
In the second phase, the collected datasets were furthered processed via (1)~\textit{Language segmentation}, whereby data sources were split using  metadata for each covered language in order to obtain monolingual datasets, and the use of (2)~\textit{Uniform interface} whereby a document consisted of two fields: "text" for the actual text content, and "meta" with a JSON representation of metadata for a given document, containing sufficient information to trace documents back to their original sources.

\paragraph{Pseudo-Crawled Data.}

Of the various categories of language resources identified through the data sourcing effort, websites stood out as one that required a particular effort and dedicated pipeline. We decided to design such a pipeline based on ``pseudo-crawling'': that is, rather than crawling the websites ourselves, we retrieved pages corresponding to the target domain names from 18 snapshots archived by Common Crawl in 2020 and 2021 in Web ARChive (WARC) format~\citep{mohr2008warc}.
These domain names came from two main sources: the homepage field in the metadata of the 252 above-mentioned catalogue entries when available (192 in total), and the 456 websites proposed by participants asynchronously to improve the geographical diversity of our language sources; which yielded a total of 614 unique domain names after deduplication.

We collected URLs contained within those domains using the Common Crawl index.
The index provides metadata for every document including the page URL, WARC filename and record offsets, fetch status, content MIME type, etc. We ran a query matching all documents that share the domain name with a seed using \href{https://docs.aws.amazon.com/athena/}{Amazon Athena} on Common Crawl's columnar index\footnote{\href{https://commoncrawl.org/2018/03/index-to-warc-files-and-urls-in-columnar-format/}{https://commoncrawl.org/2018/03/index-to-warc-files-and-urls-in-columnar-format/}}.
48 of the 614 initial seed domain names had no matches in the index and were therefore left out.
Once we obtained the document metadata, we fetched the WARC records using HTTP range requests with the start and end byte offsets. Since HTML web pages constitute the largest portion of pages contained in the Common Crawl dumps, we decided to only extract text from HTML pages. Documents in other formats were filtered out, ie XML, PDF, etc. 27 domain names were additionally removed from the list at this stage as we had not retrieved any HTML pages for them. 

To extract the text from the HTML pages, we first minified the HTML code. Minification is the removal of unnecessary characters from the source code of a website. Inspired by \citet{aghajanyan2022htlm}, we removed from the DOM-HTML all the sub-trees contained in a \textit{<script>}, \textit{<style>}, \textit{<header>}, \textit{<iframe>}, \textit{<footer>} and \textit{<form>} tag as well as all the sub-trees associated with a \textit{<body>}, \textit{<div>}, \textit{<p>}, \textit{<section>}, \textit{<table>}, \textit{<ul>}, \textit{<ol>} or \textit{<dl>} tag whose textual content was less than 64 characters long. The text was then extracted from the nodes of this new DOM-HTML. While concatenating the text extracted, we applied a set of rules to reconstruct the structure of the text without its HTML code, inspired by what Common Crawl does to extract its WET files (Appendix \ref{appendix:concatenate_extracted_texts_html}).
The overall procedure enabled us to obtain text datasets for 539 domain names.

%\footnote{When the text to be concatenated to the already concatenated texts was extracted from a tag that we consider to be a block element, if the last character of the previous text was a line break we added nothing, if it was a space we replaced it with a line break or otherwise we added a line break before concatenating the new text. When the text came from a \textit{in line} tag, if the last character was a line break we replaced it with a space, if the last character was a space we did nothing or if not we added a new space before concatenating the new text. When the text came from another type of tag we simply concatenated it to the rest.} 

% \subsection{GitHub Code}
% \label{section:obtaining-code}

% In an attempt to generate more data, we initiated a community-wide effort for people to provide "seed" URLs from which we can query all underlying pages, ie pages that share the same prefix as the "seed" URL. 456 URLs were provided from an asynchronous collection by the Data Sourcing Working Group, 192 were generated through the catalogue sprint. Both were concatenated and then deduplicated. This crowdsourcing step allowed to identify a list of 614 unique domain names. In order to retrieve the textual content contained in the websites associated with the selected domain names, we retrieved the raw web pages of these websites from the 18 snapshots archived by Common Crawl in 2020 and 2021 in Web ARChive (WARC) format.

\paragraph{GitHub Code.}

We collected a code dataset from BigQuery\footnote{\href{https://cloud.google.com/blog/topics/public-datasets/github-on-bigquery-analyze-all-the-open-source-code}{``GitHub on BigQuery: Analyze all the open source code''}} using the same language selection as AlphaCode~\citep{li2022}. The dataset was then deduplicated of exact matches and filtered for source files with between 100 and 200,000 characters, between 15-65\% alphabetic characters, a max line length of 20-1000 characters, and a token length standard deviation of more than 3. Due to a bug in the pre-processing pipeline the dataset was also filtered for GPL licenses only.

% \subsection{Merging and Further Selecting Data Sources.}

\paragraph{Merging and Deduplicating Sources.}

After gathering and processing language data via the three pipelines outlined above, we took a final step to manually inspect, deduplicate, and make a further selection of the sources. First, we addressed dataset overlap we found by looking through our sources. For example: \textit{OpenITI} was present in both its raw form as well as a processed version. Consensus was reached to choose the latter version. Non-trivial datasets overlap included \textit{s2orc} \citep{lo-wang-2020-s2orc}, \textit{Arxiv}~\citep{clement2019arxiv} and the \textit{PubMed Central} subset of the Pile \citep{Gao2020}. We also performed cross-pipeline dataset deduplication, removing the pseudo-crawled Wikipedia and GitHub in favor of their other versions. We also removed datasets that we found had a high incidence of documents that were not fully in natural language (e.g. unexpected instances of SEO, HTML tags etc...), as well as very small datasets in the higher-resourced languages. Finally, pseudo-crawled sources were further processed to remove menus (with a heuristic consisting of removing lines that occurred in more than 1\% of pages in a given domain) and pages that had a high incidence of character ngram repetition, low language identification confidence, or low proportion of closed class words (see Section~\ref{section:common_crawl_section}). We then removed entire domains whose size was less than 2MB after this step, yielding 147 pseudo-crawl-based datasets, and a total of 517 datasets including all three pipelines.
%\begin{itemize}
%    \item \textbf{Datasets overlap}: Across all the sources, datasets overlap was detected manually. For example: \textit{OpenITI} was present in both its raw form as well as a processed version. Consensus was reached to choose the latter version. Non trivial datasets overlap included \textit{s2orc} \citep{lo-wang-2020-s2orc}, \textit{Arxiv}~\citep{clement2019arxiv} and the \textit{PubMed Central} subset of the Pile \citep{Gao2020}. We also perform cross-source dataset deduplication. Typically, Wikipedia was both in primary sources and in pseudocrawl, Github was also in pseudocrawl which we manually removed.
%    \item \textbf{Automatic filtering}: TODO Yacine
%    \item \textbf{Poor quality}: Datasets were manually inspected via crowdsourced effort. Poor quality datasets from a natural language standpoint were filtered out. Criteria were to have documents that appear like natural language (no SEO, no HTML tags etc...).
%    \item \textbf{Small datasets}: In an attempt to reduce the quantity of datasets to a minimum, we filtered very small datasets (orders of magnitude smaller) if higher resources in that language existed.
%\end{itemize}

% As we processed datasets from three separate sources, via three different pipelines, we manually deduplicate the datasets as well as removed some lower quality ones:

\subsection{Processing Pipeline for Quality Improvement on Crowdsourced Datasets}

Once a text field was obtained, we attempted to improve the quality of that text. In the specific case of text extraction from HTML, we observe that not all text are relevant (menus, advertisements, repeated text on each page etc ...). In order to remove noisy data from our dataset, we applied a processing pipeline for each dataset consisting of a sequence of functions. 

Functions were categorised as \textit{document-scoped} or \textit{dataset-scoped} functions.  \textit{Document-scoped} functions are operations that modify a document independently of other documents and \textit{dataset-scoped} functions are operations that take into account the whole dataset. Orthogonal to this scope, functions were also separated into \textit{cleaning} and \textit{filtering} functions.  \textit{Cleaning functions} aim to remove text considered not part of the main document.  Document-scoped cleaning functions can for example target leftover HTML tags. On the other end, dataset-scoped cleaning functions need the whole dataset to calculate a heuristic to determine how to modify each document. For instance, advertisements vary across datasets, making it harder to define a dataset-agnostic classifier for advertisement. Instead, we can index all the lines in a dataset and identify repeated lines on multiple pages as likely advertisements. An example is displayed in Appendix~\ref{appendix:visu_tool}. \textit{Filtering functions} aim at removing an entire document from the corpus. The reasons for choosing to remove a document completely are diverse: it may be because the document is considered to be of too poor quality, to be too complex to automatically fix or too similar to other examples already present in the corpus. In the latter case, we speak of deduplication. Deduplication of a document is dependent on whether an equivalent document already exists somewhere else in the dataset and is thus necessarily a dataset-scope function. The notion of equivalent documents has been explored by \citet{deduplication-training-data-makes-language-models-better}. In this case we provide deduplication via metadata (urls, normalised urls) and via text (exact string matching).
An exhaustive list of functions is available in \ref{appendix:crowdsourced_dataset_exhaustive_list_of_functions}. 

\begin{figure}
    \centering
    \includegraphics[width=\linewidth]{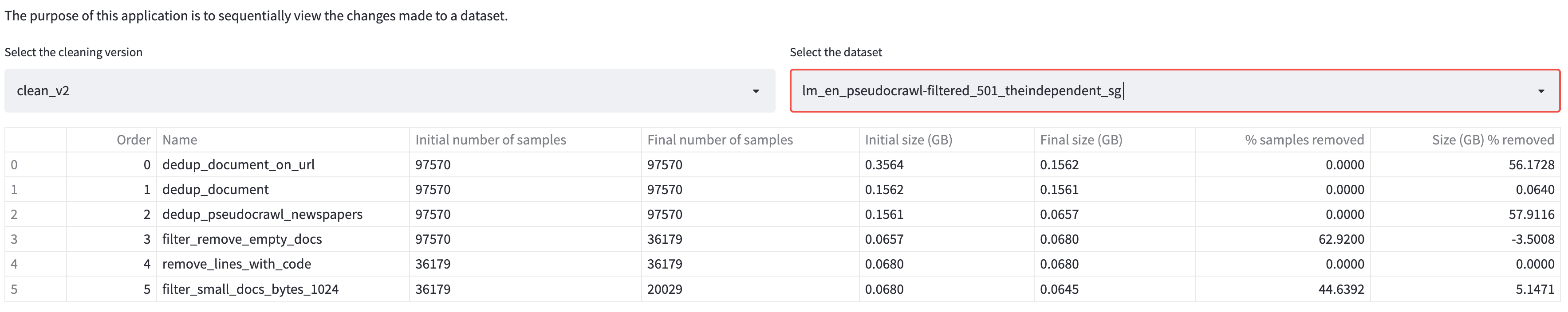}
    \caption{Partial screenshot of the visualization tool. Users can look at how each function in the processing pipeline influenced high-level statistics. Influence on specific samples can be monitored via the same tool, see Appendix~\ref{appendix:visu_tool}}
    \label{fig:catalogue_cleaning_screenshot}
\end{figure}

As datasets came from heterogeneous sources with different properties, each needs its own set of processing functions to correspond to our definition of natural language documents.
In order to support participants in deciding what functions to apply to which, we built and released a \href{https://github.com/streamlit/streamlit}{\textit{streamlit}}-based \href{https://huggingface.co/spaces/bigscience-catalogue-lm-data/process-pipeline-visualizer}{visualization tool} (figure~\ref{fig:catalogue_cleaning_screenshot} helps understand the impact of each function, displaying how a document was altered/removed as well as estimated dataset level metrics (quantity of data removed in bytes or samples)).
This rapid feedback loop enabled us to update the pipeline consequently in an iterative process to finetune each processing pipelines across datasets and languages with the input of native speakers. A specific example is shared in Appendix~\ref{appendix:visu_tool}. This resulted in 485 non-empty datasets.

\section{Processing OSCAR}
\label{section:common_crawl_section}

We chose to complement the data obtained at the end of the process described in the previous section with additional Common~Crawl-based\footnote{\href{https://commoncrawl.org/}{https://commoncrawl.org/}} data motivated by two main reasons. First, given the project's overall goal of providing a trained LLM as a research artifact comparable to previously released ones that have relied extensively on this source, we assessed that not including it would constitute too much of a departure and risk invalidating comparisons. Relatedly, recent work has put a strong emphasis on the quantity of data being a strong factor in a trained model's performance on evaluation tasks \citep{scalinglaws,chinchilla}, and we were missing about one third of data in order to optimize our compute budget in this direction. With that in mind, we chose OSCAR version 21.09 \citep{ortiz-suarez-etal-2020-monolingual}, based on the Common Crawl snapshot of February 2021, to make up the remaining 38\% of our final dataset.

However, crawled data suffers from several known issues. First, we wanted to only select documents written by humans for humans, and exclude machine-generated content e.g. search engine optimization (SEO). Crawled content also over-represents pornographic text across languages~\citep{Kreutzer2022}, especially in the form of spam ads. Finally, it contains personal information that may constitute a privacy risk. The present section outlines our approach to mitigating those issues.

% However, crawled data is often highly noisy. We wanted to remove documents not generated by humans for humans, for example, those containing machine-generated sentences, potentially built for search engine optimization (SEO), and documents with a sequence of unrelated sentences.
%Moreover, it is a known problem that language models tend to produce many repetitions during generation \citep{fu2020a}. Therefore, we also wanted to remove beforehand documents containing too many repetitions to try to reduce this effect.
% In an effort to improve diversity in documents, we also used a stronger deduplication than the exact one already done in OSCAR, to remove similar but not identical documents.
% Finally, we wanted to reduce the over-representation of pornographic material in the Common Crawl data.%, while taking care to keep a good part of the erotic texts.

\subsection{Data cleaning and filtering\label{section:cc_filters}}

\begin{figure}
    \centering
    \includegraphics[width=\textwidth]{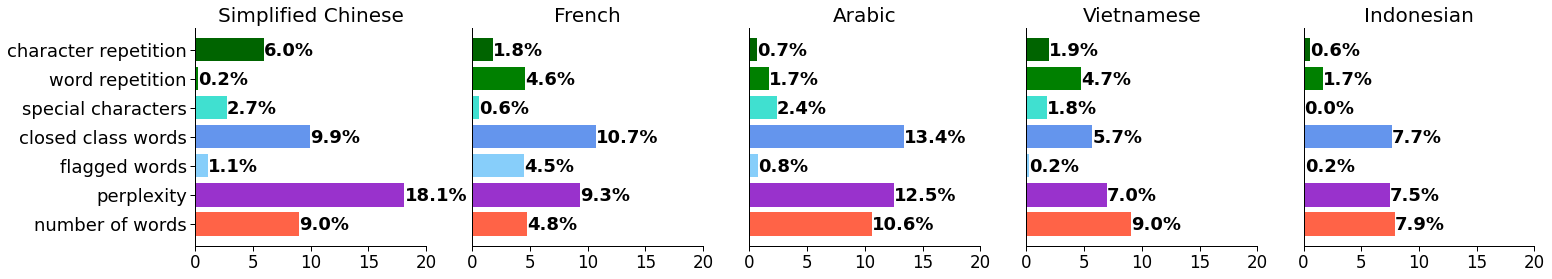}
    \caption{Percentage of documents discarded by each filter independently for 5 languages}
    \label{fig:perc_discarded_by_filter}
\end{figure}

% For this purpose, we built different filters to meet the needs encountered. These filters 
%% Thomas: remove
%, which can be activated or not,
% require in most cases the input of a cutoff threshold, which should ideally be language-dependent, and manually chosen by looking at the data.
Our first approach to addressing the above consists in defining quality indicators for web content. These can then be used to filter out specific pages by defining cutoff thresholds. Extensive descriptions for reproduction are available in appendix \ref{appendix:oscar_filters}. We filtered out documents with:
\begin{itemize}
    \item Too high \textbf{character repetition} or \textbf{word repetition} as a measure of repetitive content.
    \item Too high ratios of \textbf{special characters} to remove page code or crawling artifacts.
    \item Insufficient ratios of \textbf{closed class words} to filter out SEO pages.
    \item Too high ratios of \textbf{flagged words} to filter out pornographic spam. We asked contributors to tailor the word list in their language to this criterion (as opposed to generic terms related to sexuality) and to err on the side of high precision.
    \item Too high \textbf{perplexity} values to filter out non-natural language.
    \item Insufficient \textbf{number of words}, as LLM training requires extensive context sizes.
\end{itemize}
The languages that we eventually considered in OSCAR were the languages for which we were able to obtain hyperparameters and the cutoff values for each of these indicators by native speakers.
Specifically, we considered Arabic, Basque, Bengali, Catalan, Chinese, English, French, Hindi, Indonesian, Portuguese, Spanish, Urdu, and Vietnamese.
The code used for filtering OSCAR, along with the language-specific parameters and cutoff values, are \href{https://github.com/bigscience-workshop/data-preparation/tree/main/preprocessing/filtering}{publicly available}.
% It aims to be easily reusable for any text dataset, with the possibility to activate or not each filter or to change the cutoff values.
% The cutoff values chosen for the filtering and the deduplication steps for this project tuned for each language are also made public.
We then asked native speakers of each language to use our visualization tool\footnote{\href{https://hf.co/spaces/huggingface/text-data-filtering}{https://hf.co/spaces/huggingface/text-data-filtering}} to establish the thresholds for each filter. The percentage of documents removed after applying all these filters is given in Table~\ref{table:perc_remove_filtering}, and the percentage of documents discarded by each filter independently is given in~\ref{fig:perc_discarded_by_filter}.

\begin{table}[!ht]
\begin{center}
%\begin{tabular}{|c|c|c|c|c|c|c|c|c|c|c|c|c|} 
\begin{tabular}{|ccccccccccccc|} 
 \hline
 AR & EU & BN & CA & ZH & EN & FR & HI & ID & PT & UR & VI & ES \\ 
 \hline
 \hline
 20.3 & 5.2 & 48.8 & 21.1 & 23.1 & 17.2 & 17.0 & 25.7 & 10.4 & 12.6 & 15.8 & 21.3 & 16.9 \\ 
 \hline
\end{tabular}
\end{center}
\caption{Percentage of documents removed by the filtering per language (ISO~639-1 code).}
% [Lucile to gain 1 line] \caption{Percentage of documents removed by the filtering per language (identified by the ISO~639-1 code).}
\label{table:perc_remove_filtering}
\end{table}

% [Lucile moved to section 5]
% The distributions of the filter values for the different filters and languages, for the Catalogue, Pseudo Crawl and Common Crawl (filtered) data are available in an online demo.\footnote{\href{https://huggingface.co/spaces/HugoLaurencon/filter\_values\_distributions}{https://huggingface.co/spaces/HugoLaurencon/filter\_values\_distributions}} We did not include the filter on the language identification prediction score, since a high confidence score does not imply a correct prediction. For the flagged word ratio values, since most of them were exactly equal to 0, we removed them for visibility reasons after the density normalization. Two distributions of the filter values for the filter on the character repetition ratio and the filter on the flagged word ratio are given for English on Figure~\ref{fig:examples_plots_filter_values}.
% \begin{figure}
%     \centering
%     %\subfloat[\centering label 1]{{\includegraphics[width=6.5cm]{figures/en\_character\_repetition\_ratio.png} }}
%     \subfloat{{\includegraphics[width=6.5cm]{figures/en\_character\_repetition\_ratio.png} }}
%     \qquad
%     %\subfloat[\centering label 2]{{\includegraphics[width=6.5cm]{figures/en\_flagged\_word\_ratio.png} }}
%     \subfloat{{\includegraphics[width=6.5cm]{figures/en\_flagged\_word\_ratio_2.png} }}
%     \caption{Distributions of the filter values for the filter on the character repetition ratio (left) and the filter on the flagged word ratio (right) for English.}
%     \label{fig:examples_plots_filter_values}
% \end{figure}

\subsection{Deduplication}

Data deduplication has become a key tool for language model projects following research showing that it both improves performance on downstream tasks \citep{deduplication-training-data-makes-language-models-better,zhang2021counterfactual} and decreases memorization of training data \citep{kandpal2022deduplicating}. To remove near duplicate documents in OSCAR (which is already exact-deduplicated) we initially used SimHash \citep{10.1145/509907.509965,10.1145/1242572.1242592}, a hashing function that associates to two similar texts hashes with a low Hamming distance, with 6-grams and a Hamming distance threshold of 4. About 0.7\% of the documents on average (0.07\% $\sim$ 2.7\%) were identified as near duplicates. However, because SimHash is essentially a bag-of-words algorithm, long documents are more likely to end up being similar to each other. In practice, we found false positives among long documents and decided not to discard documents in a same cluster of near-duplicates when they were longer than 6000 characters. Instead, we applied substring deduplication \citep{deduplication-training-data-makes-language-models-better} based on Suffix Array \citep{doi:10.1137/0222058} as a complementary method that clusters documents sharing a long substring, for documents with more than 6000 characters. We found on average 21.67\% (10.61\% $\sim$ 32.30\%) of the data (in bytes) being duplicated. 
%% Thomas: Remove, as it's not something we actually used, it might be interesting, but for the sake of limited pages deciding to remove it for now.
%After the development of this corpus, further research showed that the performance loss caused by duplication is primarily due to sequences with a very high duplication rate \citep{hernandez2022scaling}, and that duplicating sequences a small number of times does not have a notable impact \citep{black2022gpt}. We were unfortunately unable to take these results into account when building our corpus, but advise that future researchers do so.

\subsection{Personally identifiable information}
\label{sec:pii_regex}

%% Thomas: Remove, not something we used to construct the corpus
% Initial experiments in PII identification and redaction from OSCAR included the creation of \texttt{muliwai}\footnote{\href{https://github.com/ontocord/muliwai}{https://github.com/ontocord/muliwai}} neural-based library which has since grown beyond the BigScience context. The scale of the data and the impact on the resulting text could not be fully assessed meant this approach could not be operationalized.
We  used a rule-based approach leveraging regular expressions (Appendix~\ref{appendix:oscar_filters}). The elements redacted were instances of \textit{KEY} (numeric \& alphanumeric identifiers such as phone numbers, credit card numbers, hexadecimal hashes and the like, while skipping instances of years and simple numbers), \textit{EMAIL} (email addresses), \textit{USER} (a social media handle) and \textit{IP\_ADDRESS} (an IPv4 or IPv6 address).

% \subsection{Data selection and methodology to improve data quality}
% In order to provide sufficient diversity and quantity of data, it was necessary to include texts from non-manually selected sources.
%We  chose OSCAR \citep{ortiz-suarez-etal-2020-monolingual}, a huge multilingual corpus obtained by language classification and filtering of the Common Crawl\footnote{\href{https://commoncrawl.org/}{https://commoncrawl.org}} corpus. We used OSCAR version 21.09, based on the Common Crawl snapshot of February 2021, which spans a total of 168 languages. An exact deduplication across documents was also performed for each language.

% Postprocessing: Tokenization fertility, filtering, PII -> THIS WAS MERGED INTO SECTIONS 3 AND 4
% \input{section_5}

% A First look at the Dataset
\section{A First look at ROOTS}
\label{section:first-look}
\begin{figure}[t]
\centering
\includegraphics[width=0.6\textwidth]{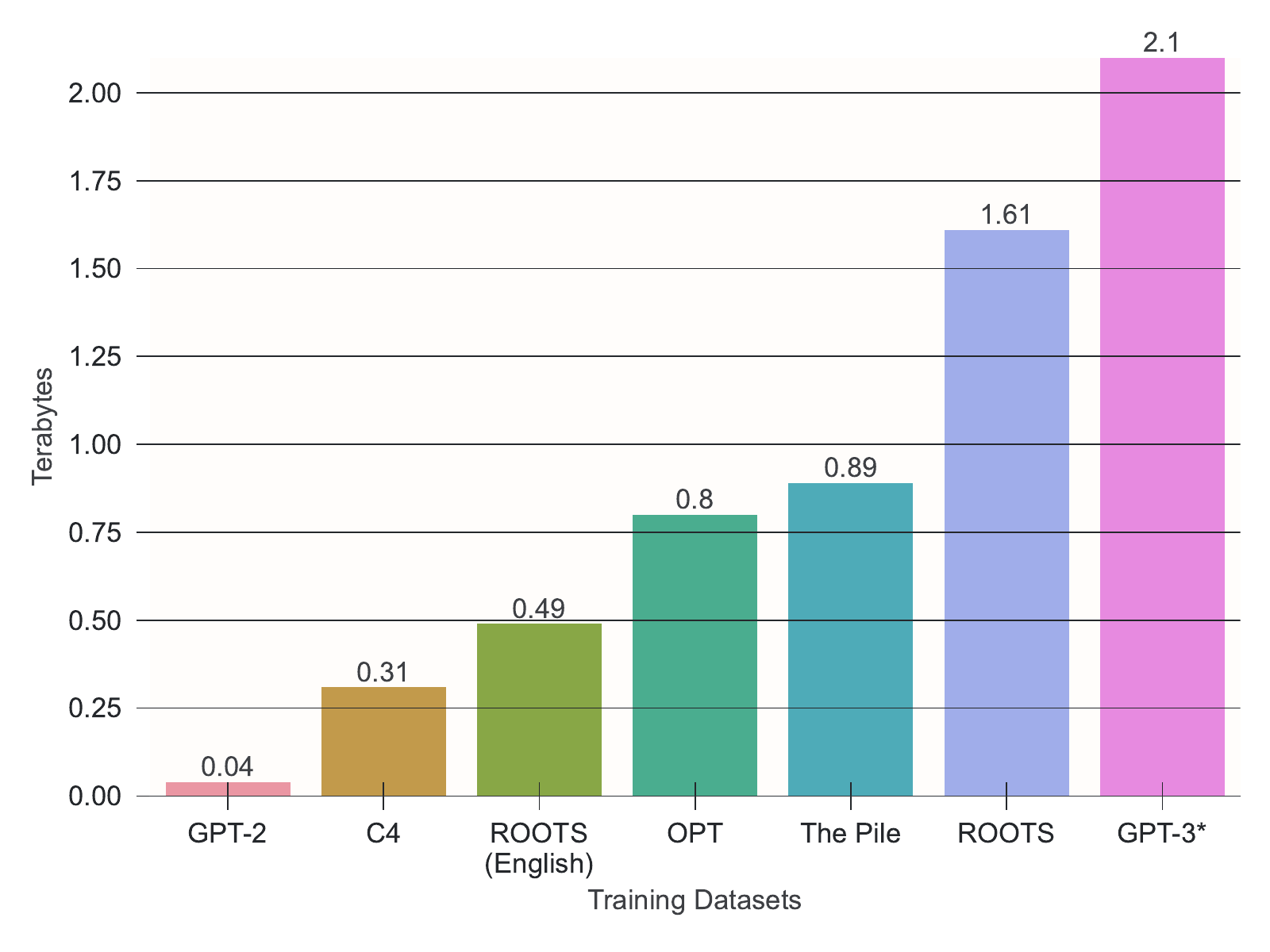}
\caption{A raw size comparison to other corpora used to train large language models. The asterisk next to GPT-3 indicates the fact that the value in question is an estimate computed using the reported number of tokens and the average number of tokens per byte of text that the GPT-2 tokenizer produces on the \texttt{Pile-CC}, \texttt{Books3}, \texttt{OWT2}, and \texttt{Wiki-en} subsets of the Pile~\citep{Gao2020}}
\label{fig:comparison}
\end{figure}

The efforts described in the previous sections come together in an assemblage of 1.6 Terabytes of multilingual text. Figure~\ref{fig:comparison} puts that number into context by comparing the sizes of corpora typically used to train large language models.
Documentation of the individual components of the corpus can be found in an interactive \href{https://hf.co/spaces/bigscience/BigScienceCorpus}{dataset card deck}.
In this section, we take initial steps towards further understanding of the corpus through statistical analyses of the aggregated data.

\subsection{Natural Languages}
The constitution of the corpus reflects the crowdsourcing efforts that enabled its creation. It comprises of 46 natural languages spanning 3 macroareas and 9 language families: Afro-Asiatic, Austro-Asiatic, Austronesian, Basque, Dravidian, Indo-European, Mande, Niger-Congo, Sino-Tibetan. At 30.03\%, English constitutes the largest part of the corpus, followed by Simplified Chinese~(16.16\%), French~(12.9\%), Spanish~(10.85\%), Portuguese~(4.91\%) and Arabic~(4.6\%). A more detailed breakdown of the corpus can be found in the appendix and in an online interactive exploration tool\footnote{\href{https://hf.co/spaces/bigscience-catalogue-lm-data/corpus-map}{https://hf.co/spaces/bigscience-data/corpus-map}}, a screenshot of which is included in figure~\ref{fig:corpus_distribution} to depict the byte-distribution of linguistic genera of the Eurasian macroarea subset of the corpus.

\begin{figure}[t]
\includegraphics[width=\textwidth]{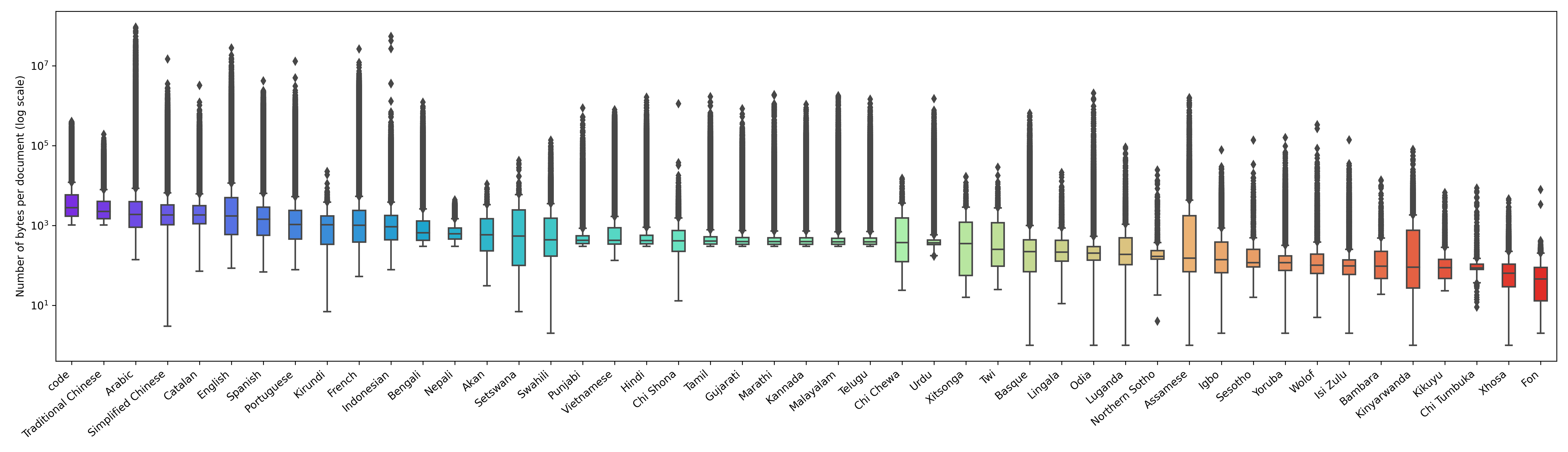}
\caption{Size in bytes of every document in the corpus per language. The y-axis is in logarithmic scale. Box-and-whisker diagrams illustrate median, the first and third quartiles, whiskers drawn within the 1.5 IQR value and outliers}
\label{fig:dataset_document_sizes}
\end{figure}

In order for the trained model to have an opportunity to learn long dependencies, the training corpus needs to contain long sequences of coherent text.
At the same time, the previous post-processing steps only reduced the size of the documents. The median size of a document in our corpus is 1,129 bytes.
Figure~\ref{fig:dataset_document_sizes} shows the distribution of document sizes by language. A more detailed breakdown of the size of corpus on an online interactive tool.\footnote{\href{https://hf.co/spaces/bigscience-catalogue-lm-data/document-sizes}{https://hf.co/spaces/bigscience-data/document-sizes}}.   

\begin{figure}[t]
\includegraphics[width=\textwidth]{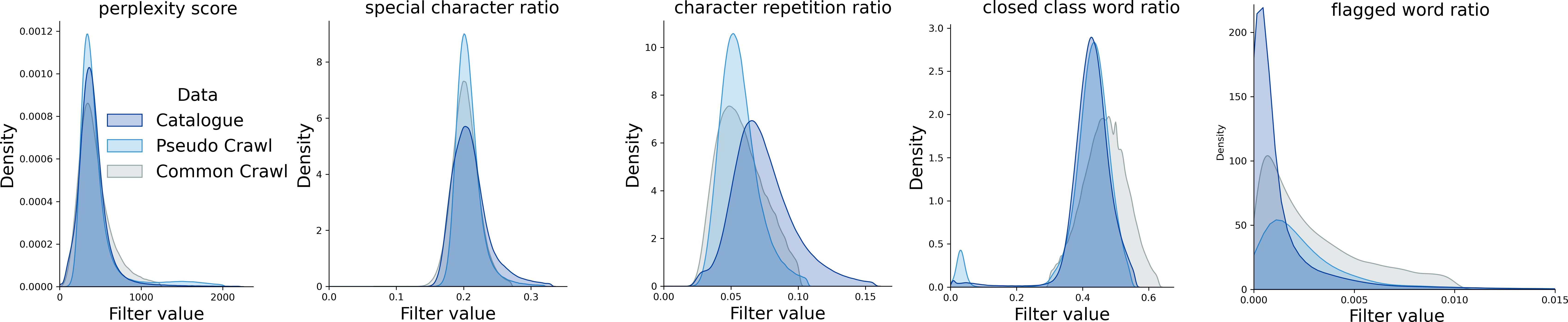}
\caption{Some distributions of filter values for English. A filter value is the value that the filter gives to a document. These values are generally used to filter out documents that are too low or too high rated and also inform about the composition of the datasets.}
\label{fig:examples_plots_filter_values}
\end{figure}

The distributions of the filter values for the different filters introduced in Section \ref{section:cc_filters} and languages, for the Catalogue, Pseudo-Crawl and OSCAR (filtered) data are available in an online demo\footnote{\href{https://hf.co/spaces/bigscience-data/filter\_values\_distributions}{https://hf.co/spaces/bigscience-catalogue-lm-data/filter\_values\_distributions}}. Examples for English are shown in figure~\ref{fig:examples_plots_filter_values}. The different distributions reflect the diversity of sourcing and filtering of our main components. A notable example is the flagged word filter, for which the distribution for OSCAR is skewed right compared to the catalogue even after filtering.

\subsection{Programming Languages} 
As depicted in the waffle plot in figure~\ref{fig:corpus_distribution}, the code subset of the corpus spans 13 programming languages, with Java, PHP, and C++ accounting for more than half of all documents.

Configuration and test files are abundant in most GitHub repositories but not as interesting for code modeling. To that end, we use a heuristic whose first step examines the first 5 lines of a file for the presence of keywords such as ``configuration file'' or ``test file''. Failing that, the second step is to see whether the occurrence of the literals \texttt{config} and \texttt{test} in a given file exceeds 5\% of the total number of lines of that file. We find that 5.23\% of the data consists of configuration files and 7.88\% of test files.

\citet{allamanis2019adverse} and \citet{lopes2017dejavu} highlight the large fraction of near-duplicates present in code datasets and how they can inflate performance metrics. Exact match deduplication alone can miss a fair amount of near-duplicates. To detect them, we first compute the MinHash of all documents, then create a Locality Sensitive Hashing (LSH) index between files to find the duplicate clusters in linear time. We additionally evaluate the Jaccard similarities within duplicate clusters to remove some false positives. We find 10.9M duplicate files in the clusters and 4.1M unique files: almost 32\% of the data consists of near-duplicates. Syntax checkers\footnote{\texttt{py\_compile} for Python and the \texttt{-l} flag for PHP} are used to validate 500K samples of Python and PHP code. We find that only 1\% of the Python data and 2\% of the PHP files do not pass the syntax check.

% In order for trained model to display long dependencies in long sequences, a key aspect is to have documents that are long and that have long dependencies underneath. 
% \begin{figure}[t]
% \includegraphics[width=\textwidth]{figures/boxplot_sizes_fr_datasets.jpeg}
% \caption{Size in bytes of every document in the French subset of the corpus. The y-axis is in logarithmic scale.}
% \label{fig:boxplot_fr}
% \end{figure}

\subsection{Tokenizer analysis of the component datasets}

A tokenizer trained on a dataset can be used as a proxy for its content \citep{Gao2020}. The relevant metric is the number of tokens produced for a byte of natural language. The more different the training corpus from the tokenized corpus, the more tokens will be produced as the tokenizer is forced to divide natural text in more numerous, more general, smaller tokens. This property has allowed us to spot errors associated with outlier values, such as incorrectly classified languages, or crawling error. In the following analysis, we use it in two ways: first, we can use tokenizers trained on different corpora to see how ours differs from them; and second, we can use a tokenizer trained on this corpus to assess which components are outliers. We exclude outliers smaller than 5 documents.

\begin{figure}[t]
\includegraphics[width=\textwidth]{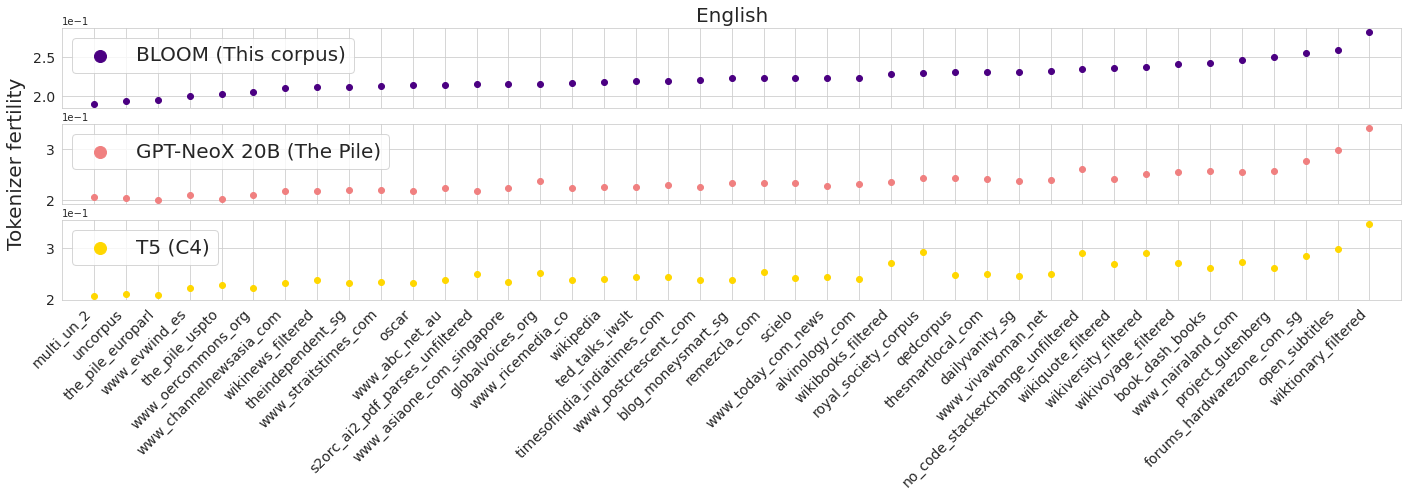}
\caption{Tokens per byte for each English-language component for tokenizers trained on this corpus (BLOOM), the Pile (GPT-NeoX 20B) and C4 (T5). Lower values mean the component (X axis) is more similar in aggregate to the compared training corpus.}
\label{fig:comparison_with_other_tokenizers}
\end{figure}

\begin{figure}[t]
\includegraphics[width=\textwidth]{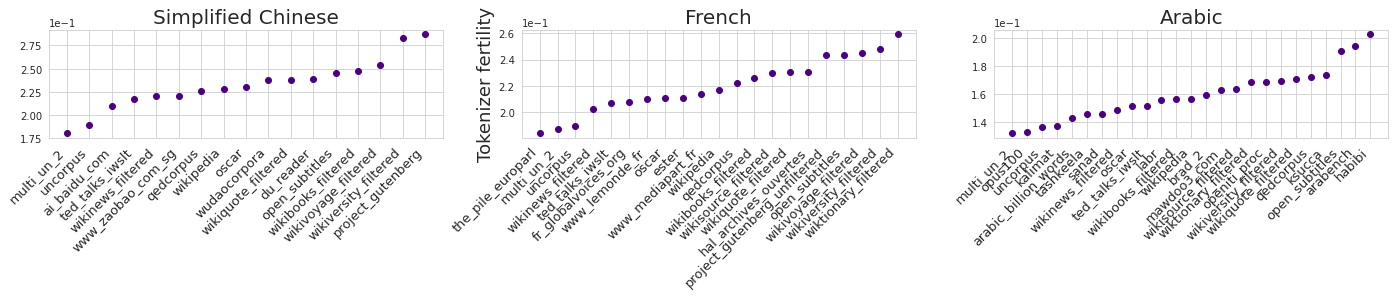}
\caption{Tokens per byte for each French, Simplified Chinese, and Arabic component for tokenizers trained on this corpus. Lower values mean the component (X axis) is more similar in aggregate to the rest of the corpus.}
\label{fig:comparison_per_language}
\end{figure}

Figure~\ref{fig:comparison_with_other_tokenizers} shows the tokens-per-byte measurement on English component datasets for the BLOOM tokenizer, trained on this corpus, the GPT-NeoX 20B tokenizer \citep{black2022gpt}, trained on the Pile, and the T5 tokenizer \citep{raffel2020exploring}, trained on C4. Those tokenizers may differ in algorithms and/or vocabulary size, but we won't be directly comparing them to each other.

The figure is ordered by BLOOM tokenizer token-per-byte values, which shows that the ordering is very similar for BLOOM and GPT-NeoX. However, it shows several bumps for T5: component datasets that are out of domain in C4 but not our corpus, for example technical and academic datasets such as \texttt{s2orc} or \texttt{royal\_society\_corpus}, domains absent from C4's Common Crawl-sourced data. Other such datasets include \texttt{global\_voices}, which contains news about non-English-speaking regions including quotes in the original languages and \texttt{no\_code\_stackexchange}, which contains forums which, although in English, may be dedicated to technical matters, foreign languages, or very specific domains. Both are similar to our corpus but not to the Pile or C4.

Figure \ref{fig:comparison_per_language} additionally shows BLOOM fertilities for Simplified Chinese, French and Arabic components. Outlier, high-fertility components, e.g. datasets that differ from the rest of our corpus, tend to be the same for all languages. \texttt{project\_gutenberg} contains old books with their original formatting (for example, "***********" to denote page ends). \texttt{wiktionary} contains definitions of words in foreign languages. \texttt{wikiversity} contains technical terms and \LaTeX. \texttt{wikivoyage} contains tables formatted as text. Forums may contain the user and date information of the message, as well as internet slang or emoji. \texttt{arabench} is spoken Arabic, and \texttt{habibi} is classical Arabic with more diacritics than modern. We deem most of those deviations acceptable to represent the diversity of uses of text, which tokenizer analysis is able to surface from the rest of the dataset.
\section{Conclusion}
We have presented ROOTS, a massive multilingual corpus that was the result of an international collaboration between multidisciplinary researchers studying large language models. The efforts to put the corpus together were value-driven and prompted by a data-first approach to training the BLOOM model. We further release the tooling developed throughout the project, and are currently implementing a release strategy that is informed by both the licensing and governance needs of every data source for the corpus itself. We hope this paves the way toward a more reflected use of the data that makes its way into large language models.
\section*{Ethical Considerations and Broader Impacts Statement}\label{appendix:ethics_main_paper}

As discussed in Section \ref{sec:intro}, the BigScience Research Workshop was conceived as a collaborative and value-driven endeavor from the start. This approach shaped many of the decisions described in this paper, spurring many contextual discussions and consensus-seeking on how to articulate the project's core values, those of the contributors to the data efforts, and considerations of social impact on the people directly and indirectly impacted. Of particular relevance were the data release and governance strategy, the choice to center human selection of data while still using OSCAR web-crawled for a significant section of the corpus, and the tools we developed to manage the risks of the latter (including regarding privacy). Each of these were the occasion of moral exercises and technical contributions that we believe were useful and required, and each will require further research and progress. We provide a more detailed discussion of these aspects of our work in Appendix~\ref{appendix:ethics}.

\section*{Acknowledgements}

\textbf{BigScience.} This work was pursued as part of the BigScience research workshop, an effort to collaboratively build a very large multilingual neural network language model and a very large multilingual text dataset. This effort gathered 1000+ reasearchers from 60 countries and from more than 250 institutions.

\textbf{Compute.} The BigScience Workshop was granted access to the HPC resources of the Institut du développement et des ressources en informatique scientifique (IDRIS) du Centre national de la recherche scientifique (CNRS) under the allocation 2021-A0101012475 made by Grand équipement national de calcul intensif (GENCI). Model training ran on the Jean-Zay cluster of IDRIS, and we thank the IDRIS team for their responsive support throughout the project, in particular Rémi Lacroix.

\bibliographystyle{chicago}
\bibliography{bibliography}

%%%%%%%%%%%%%%%%%%%%%%%%%%%%%%%%%%%%%%%%%%%%%%%%%%%%%%%%%%%%
%\input{extras/checklist}
%%%%%%%%%%%%%%%%%%%%%%%%%%%%%%%%%%%%%%%%%%%%%%%%%%%%%%%%%%%%

\newpage

\appendix
\section*{Appendix}

%\hrulefill\par
%\localtableofcontents
%\hrulefill\par

\newcommand\DoToC{%
  \startcontents
  \printcontents{}{1}{\textbf{Contents}\vskip3pt\hrule\vskip5pt}
  \vskip3pt\hrule\vskip5pt
}

%\DoToC

\section{Ethical Considerations and Broader Impacts Statement}\label{appendix:ethics}

As discussed in Section \ref{sec:intro}, the BigScience Research Workshop was conceived as a collaborative and value-driven endeavor from the start. All the ethical efforts were concentrated on implementing the values chosen first on the ethical charter and then on how to articulate those core values into specific ethical sensitive issues, such as data governance. This mechanism also allows ethical thinking to guide governance regarding technical matters. The articulation between the BigScience core values and those chosen by the collaborators contributing to data efforts was central. The importance of this collective exercise is due to the social impact that technologies such as LLMs have on the people impacted, directly and indirectly, positively and negatively. Moral exercises based on consensus, discussion around values, and how to link technical actions to ethical reflections is a strength that we believe is important within ML research. A critical analysis from an ethical perspective is fundamental to making different disciplines coexist in thinking around the social impact of these technologies and well define the object of analysis, as in this case, a multilingual dataset.

\subsection*{BigScience Values}

Motivated by recent work on the values encoded in current approaches to research in NLP and ML more broadly \citep{leahy2021hard,Birhane2021Values}, which finds that narrow definitions of performance and efficiency were often prioritized over considerations of social impact in research and development. Even more relevant to the corpus creation aspect of our project, \cite{Scheuerman2021} outline how data efforts in computer vision tend to prioritize ``\textit{efficiency [over] care; universality [over] contextuality; impartiality [over] positionality}\ldots''. These ML research programs and systems in turn support the development of new technologies that carry these same values when deploying these technologies in production~\citep{Winner2017DoAH}. This limits the potential positive societal benefits of the rapid advances of NLP research while increasing risks considerably.

Aware of these challenges, participants in BigScience collaboratively drafted an ethical charter\footref{charter} formalizing our core values and how they are articulated. It establishes the core values in order to allow its contributors to commit to them, both individually and collectively, and to ground discussions and choices made throughout the project in a common document. These values include notably \textbf{openness} and \textbf{reproducibility} as a scientific endeavor aimed at advancing the state of the art in a way that can be understood, interrogated, and re-used; \textbf{responsibility} of the participants to consider the social and legal context, and the social and environmental consequences of their work; and \textbf{diversity} and \textbf{inclusivity}. These last two are especially relevant to our data efforts, which aim to include text representative of diverse languages, varieties, and uses through a participatory approach to curation.

\subsection*{Putting Our Values into Practice}

\paragraph{Centering Participation in Data Curation}
Participatory approaches play a vital role in bridging the gaps between model development and deployment and in promoting fairness in ML applications~\citep{participatory-ml-health-fairness}. They have received increased attention in recent years, with newer work calling to involve participants as full stake-holders of the entire research life-cycle rather to catering their role to \textit{post hoc} model evaluation~\citep{sloane2020participation,guiding-principles-participatory-nlp,envisioning-communities}, as exemplified by an organization like Maskhane \citep{masakhane-participatory} that brings together African researchers to collaboratively build NLP for African languages.

With regard to developing LLMs, BigScience stands in contrast to previous work on models of similar size \citep{brown2020,zhang2022opt} --- where the majority of the development occurs in-house --- by promoting engagement with other communities at every stage of the project from its design to the data curation to the eventual model training and release. Specifically, on the data curation aspect which is the focus of this paper, the involvement of a wide range of participants from various linguistic communities aims to help with the following aspects. First, \citet{Kreutzer2022} have shown in recent work that multilingual text data curation done without involving language-specific expertise leads to resources that are very different from the intentions of their creators, and these limitations carry on to the models trained on these datasets. Second, resources that are developed in collaboration with other communities are more likely to be more directly relevant to them, and thus to avoid reduce replication of model development by making the artifacts and tools we develop useful to more people and for more languages. Third, intentional curation and proper documentation of web-scale corpora takes a significant amount of human work and expertise, which can be distributed between a large number of participants in community efforts. Finally, community involvement can help foster trust and collective ownership of the artifacts we create.

\paragraph{Addressing the Legal Landscape}
The legal status of webscraped datasets is extremely unclear in many jurisdictions, putting a substantial burden on both data creators and data users who wish to be involved with this process. While the principle of fair use generally protects academic researchers, it is not recognized in all jurisdictions and may not cover research carried out in an industry context. In consultation with our \textbf{Legal Scholarship} and \textbf{Data Governance} working groups, we developed a framework \citep{jernite2022governance} to uphold the rights and responsibilities of the many stakeholders in NLP data generation and collection, and provide assurances to downstream users as to how they are and are not authorized to use the dataset \citep{contractor2020behavioral}.

\subsection*{Limitations of the Approach.}
While we believe that an approach grounded in community participation and prioritizing language expertise constitutes a promising step toward more responsible data curation and documentation, it still has important limitations. Among those, we primarily identify the use of data from the Common Crawl which represents a point of tension between our drive to present a research artifact that is comparable to previous work and values of consent and privacy (see Section~\ref{section:common_crawl_section}). Our pre-processing removes some categories of PII but is still far from exhaustive, and the nature of crawled datasets makes it next to impossible to identify individual contributors and ask for their consent. Similar concerns apply to other existing NLP datasets we identified in the catalogue, including notably the WuDao web-based corpus \citep{yuan2021wudaocorpora} which makes up a significant part of the Chinese language data. Additionally, while we hope that our intentional approach to selecting diverse data sources (mostly along axes of geographical diversity and domains) will lead to a more representative language dataset overall, our reliance on medium to large sources of digitized content still over-represents privileged voices and language varieties.

\section{Details on tools used to obtain crowdsourced dataset}

\subsection{Pseudocode to recreate the text structure from the HTML code}\label{appendix:concatenate_extracted_texts_html}
\begin{figure}[ht]
    \centering
    \subfloat[HTML code]{
        \includegraphics[width=\textwidth]{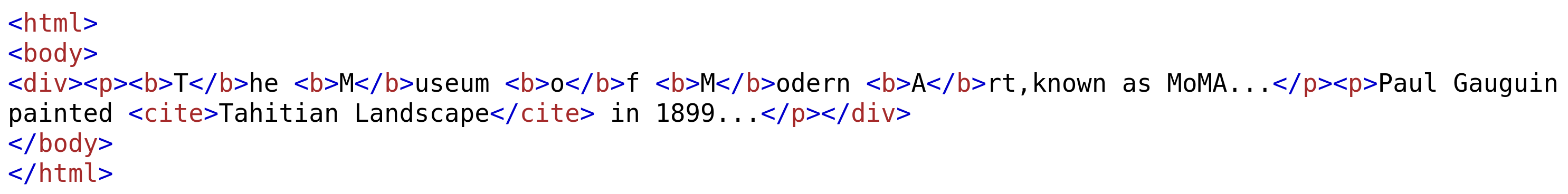}
        \label{fig:html_code}
    }
    \qquad
    \subfloat[Web browser rendering]{
        \includegraphics[width=\textwidth]{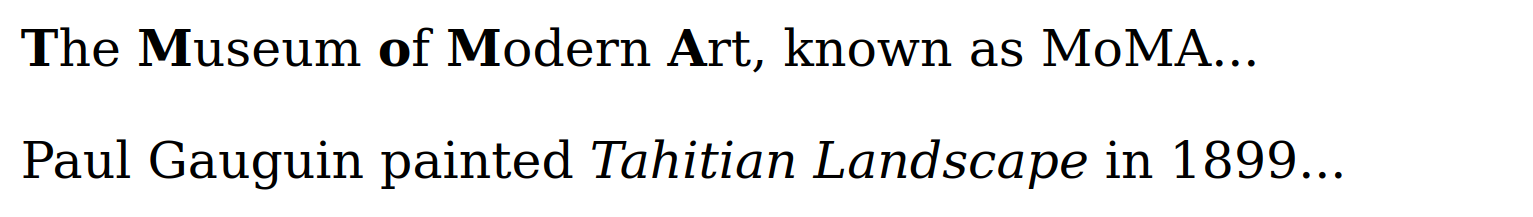}
        \label{fig:rendering}
    }
    \caption{Example showing how a single line of HTML code is rendered by a browser's renderer. In this example, we can see that the tags \textit{<p>} delimit different blocks which are therefore spaced by line breaks while other tags, such as \textit{<cite>}, are rendered on the same line of text that precedes and follows them.}
    \label{fig:structure_coming_from_rendering}
\end{figure}

\begin{figure}[htbp]
\begin{tabular}{p{0.45\textwidth}p{0.45\textwidth}}
    \begin{minipage}{.45\textwidth}
\begin{lstlisting}[style=htmlcssjs]
<div>
    <h1>Heading</h1>
    <p>
        p-inner
    </p>
    p-trailing
</div>
        \end{lstlisting}
\end{minipage}
&
    \begin{minipage}{.45\textwidth}
    \centering
    \includegraphics[width=\linewidth]{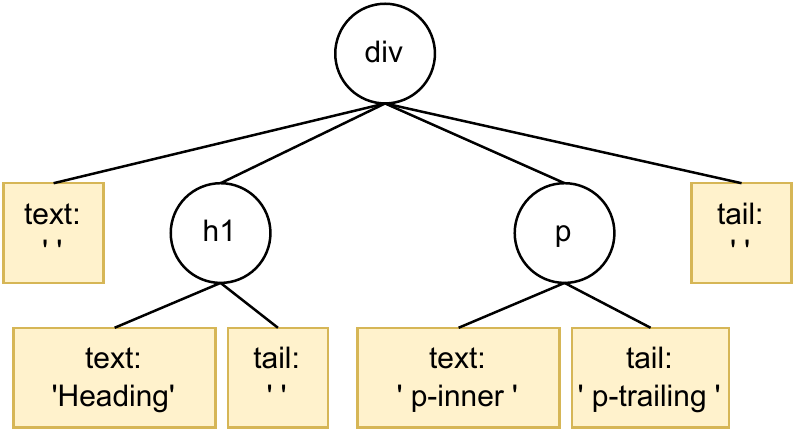}
    \end{minipage}
\end{tabular}
    \caption{Simplified version of HTML DOM model on an example. Left: snippet of HTML code. Right: corresponding DOM. The yellow squares represent nodes with textual content.}
    \label{fig:example_dom_html}
\end{figure}

The HTML code of a web page provides information about the structure of the text. The final structure of a web page is, however, the one produced by the rendering engine of the web browser and any CSS instructions. The latter two elements, which can vary enormously from one situation to another, always use the tag types for their rendering rules (Figure ~\ref{fig:structure_coming_from_rendering}. Therefore, we have used a fairly simple heuristic on tag types to reconstruct the structure of the text extracted from an HTML code. To reconstruct the text, the HTML DOM, which can be represented as a tree (Figure ~\ref{fig:example_dom_html}), is traversed with an depth-first search algorithm. The text is initially empty and each time a new node with textual content is reached its content is concatenated according to the rules presented in the Algorithm ~\ref{alg:concatenate_texts}. Block-type tags are for us: \textit{<address>},
\textit{<article>},
\textit{<aside>},
\textit{<blockquote>},
\textit{<body>},
\textit{<br>},
\textit{<button>},
\textit{<canvas>},
\textit{<caption>},
\textit{<col>},
\textit{<colgroup>},
\textit{<dd>},
\textit{<div>},
\textit{<dl>},
\textit{<dt>},
\textit{<embed>},
\textit{<fieldset>},
\textit{<figcaption>},
\textit{<figure>},
\textit{<footer>},
\textit{<form>},
\textit{<h1>},
\textit{<h2>},
\textit{<h3>},
\textit{<h4>},
\textit{<h5>},
\textit{<h6>},
\textit{<header>},
\textit{<hgroup>},
\textit{<hr>},
\textit{<li>},
\textit{<map>},
\textit{<noscript>},
\textit{<object>},
\textit{<ol>},
\textit{<output>},
\textit{<p>},
\textit{<pre>},
\textit{<progress>},
\textit{<section>},
\textit{<table>},
\textit{<tbody>},
\textit{<textarea>},
\textit{<tfoot>},
\textit{<th>},
\textit{<thead>},
\textit{<tr>},
\textit{<ul>}, and 
\textit{<video>}. Inline-type tags are for us: \textit{<address>},
\textit{<cite>},
\textit{<details>},
\textit{<datalist>},
\textit{<iframe>},
\textit{<img>},
\textit{<input>},
\textit{<label>},
\textit{<legend>},
\textit{<optgroup>},
\textit{<q>},
\textit{<select>},
\textit{<summary>},
\textit{<tbody>},
\textit{<td>}, and
\textit{<time>}.

%\footnote{When the text to be concatenated to the already concatenated texts was extracted from a tag that we consider to be a block element, if the last character of the previous text was a line break we added nothing, if it was a space we replaced it with a line break or otherwise we added a line break before concatenating the new text. When the text came from a \textit{in line} tag, if the last character was a line break we replaced it with a space, if the last character was a space we did nothing or if not we added a new space before concatenating the new text. When the text came from another type of tag we simply concatenated it to the rest.} 
% Lucile's version
\begin{algorithm}
  \caption{Pseudo-code to concatenate texts retrieved from the HTML DOM}\label{alg:concatenate_texts}
  \begin{algorithmic}[1]
    \State $text$ $\gets$ empty string
      \For{$new\_text$ \textbf{in} \texttt{list of texts retrieved by the DFS traversal}}
        
        \If{$new\_text$ is attached to a block-type tag} 
        \State \textit{\# \texttt{Block elements are separated from the rest by a line break}}
            \If{$text$ ends with a breaking line}
            \State $text$ $\gets$ $text$ + $new\_text$
            \ElsIf{$text$ ends with a space}
            \State $text$ $\gets$ $text$ without end space
            \State $text$ $\gets$ $text$ + breaking line + $new\_text$
            \Else
            \State $text$ $\gets$ $text$ + breaking line + $new\_text$
            \EndIf
        \ElsIf{$new\_text$ is attached to a inline-type tag}
            \State \textit{\# \texttt{Inline elements are separated from the rest by a line break or a space}}
            \If{$text$ ends with a space or a breaking line}
            \State $text$ $\gets$ $text$ + $new\_text$
            \Else
            \State $text$ $\gets$ $text$ + space + $new\_text$
            \EndIf
        \Else
        \State $text$ $\gets$ $text$ + $new\_text$
        \EndIf
      \EndFor
  \end{algorithmic}
\end{algorithm}
% % Thomas' version
% \begin{algorithm}
%   \caption{Pseudo-code to concatenate texts retrieved from the HTML DOM}\label{alg:concatenate_texts}
%   \begin{algorithmic}[1]
%     \State $text$ $\gets$ empty string
%       \For{$new\_text$ \textbf{in} \texttt{list of texts retrieved by the DFS traversal}}
%         \If{$new\_text$ is attached to a block-type tag}
%             \If{$text$ ends with a space}
%                 \State $text$ $\gets$ $text$ without end space + breaking line
%                 \Comment{Trim accumulator}
%             % \EndIf
        
%             \ElsIf{$text$ does not end with a breaking line}
%                 \State $text$ $\gets$ $text$ + breaking line
%                 \Comment{Add breaking line}
%             \EndIf
%         \ElsIf{$new\_text$ is attached to a inline-type tag}
%             \If{$text$ does not end with a space or a breaking line}
%                 \State $text$ $\gets$ $text$ + space
%                 \Comment{Add space}
%             \EndIf
%         \EndIf
%         \State $text$ $\gets$ $text$ + $new\_text$
%         \Comment{Concatenate new text in the accumulator}
%       \EndFor
%   \end{algorithmic}
% \end{algorithm}

\subsection{Visualization tool use cases}
\label{appendix:visu_tool}

\begin{figure}
    \centering
    \includegraphics[width=\linewidth]{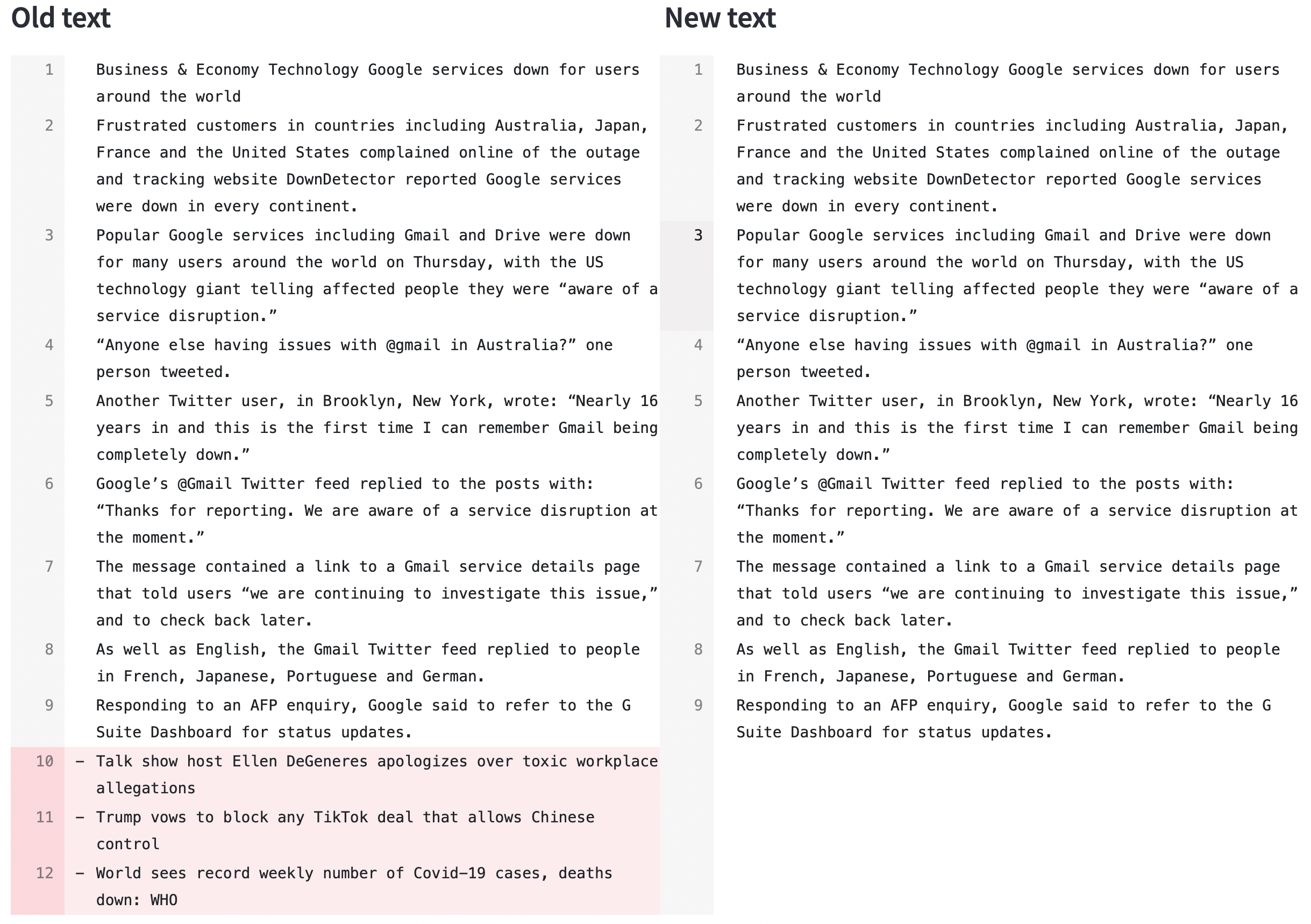} 
    \caption{Example of showing sample changes throughout each step of the processing pipeline. In the following example, users can notice that advertisement text were removed from the main article.}
    \label{fig:catalogue_cleaning_screenshot_sample}
\end{figure}

The visualisation tool was for us an iterative tool that we used to define new cleaning and filtering methods by visualising their effect on a subset of documents. This visualisation allowed us understand the impact of functions on the dataset at every stage of the processing pipeline (Figure \ref{fig:catalogue_cleaning_screenshot_sample} for advertisement detection for example), prompted us to adapt pipelines as well as introduce new functions for specific cases.

\begin{figure}
    \centering
    \subfloat[Pipeline v0]{
        \includegraphics[width=\textwidth]{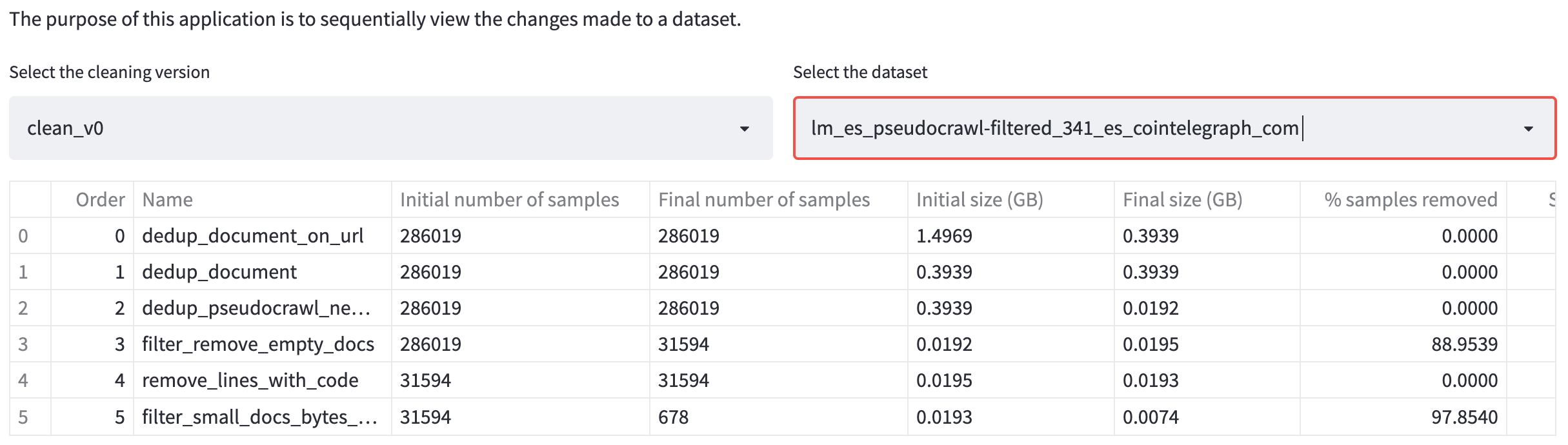}
        \label{fig:pipeline_v0}
    }
    \qquad
    \subfloat[Pipeline v2]{
        \includegraphics[width=\textwidth]{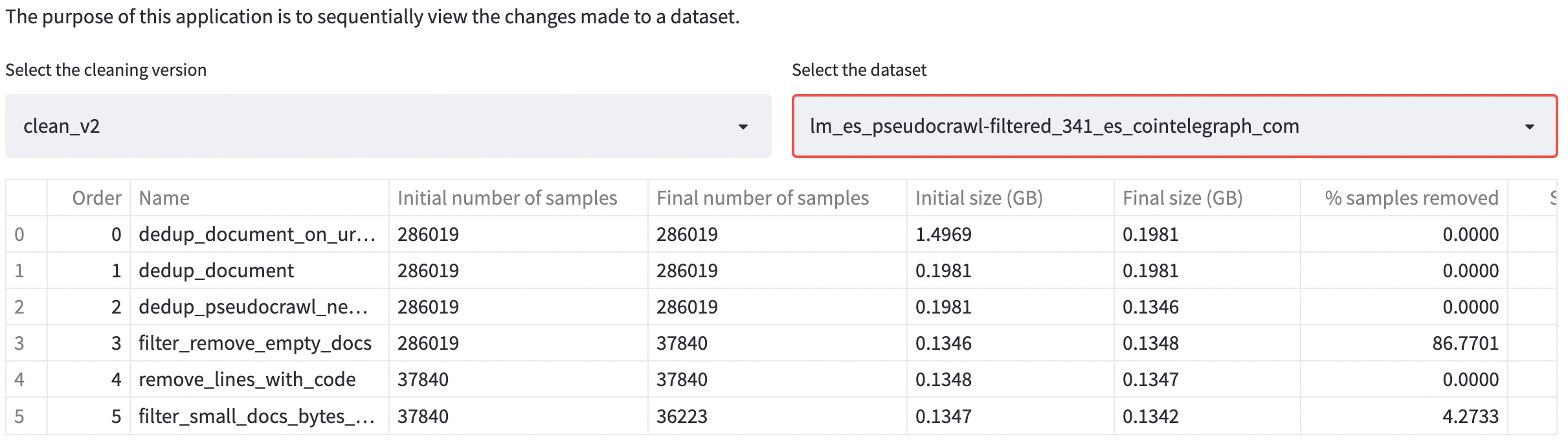}
        \label{fig:pipeline_v2}
    }
    \qquad

    \subfloat[Sample example difference between pipeline versions]{
        \includegraphics[width=0.49\textwidth]{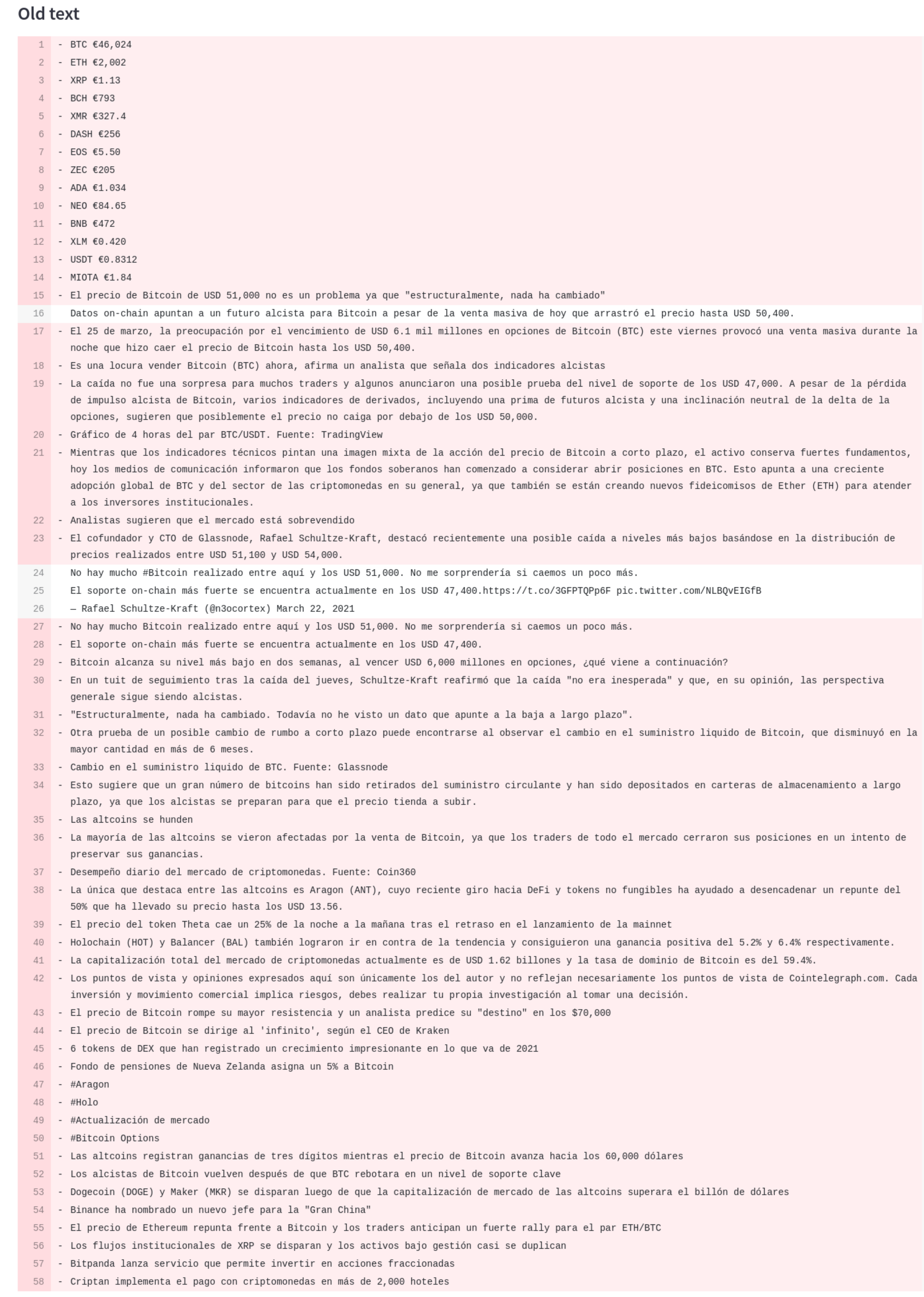}
        \hfill
        \includegraphics[width=0.49\textwidth]{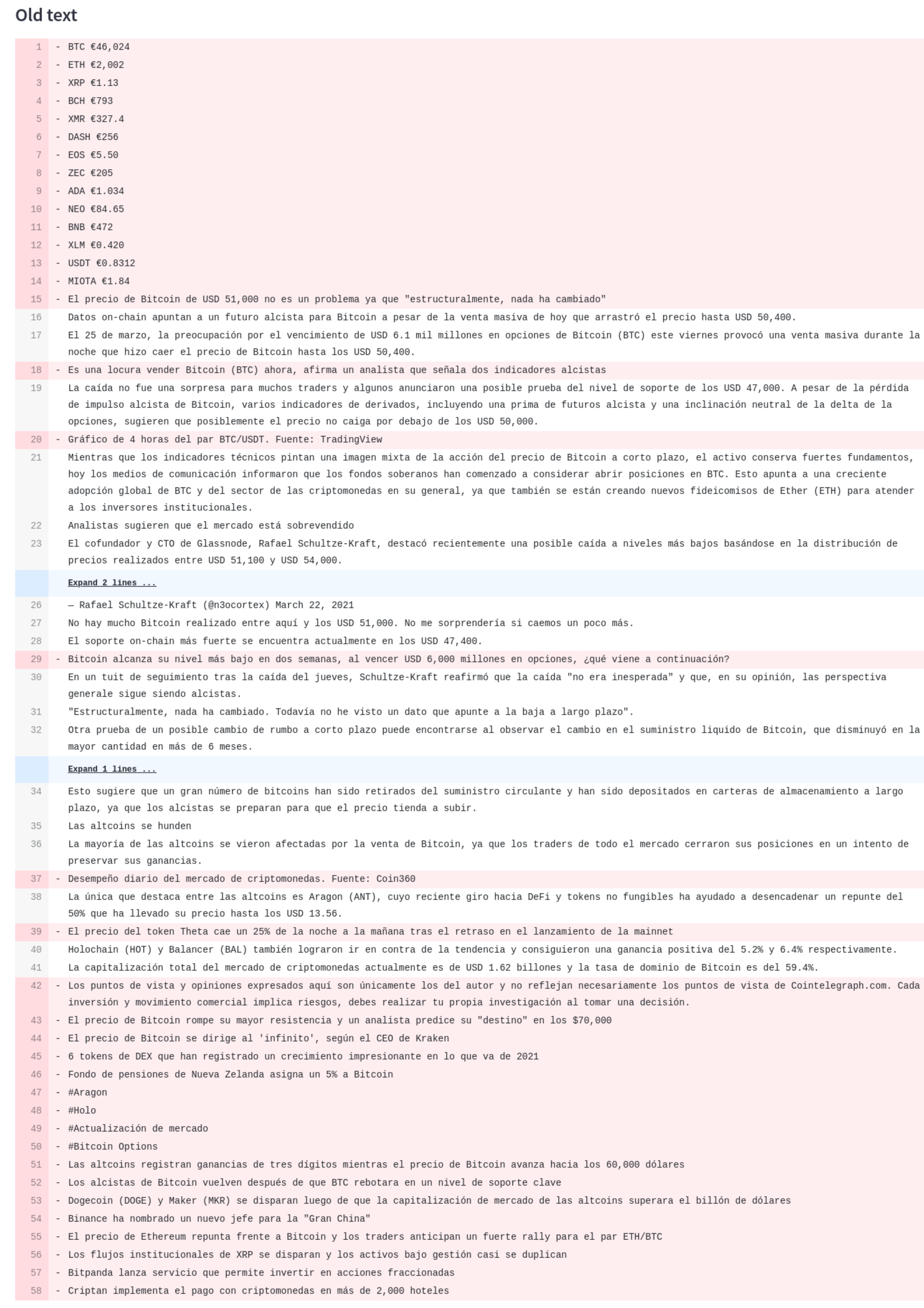}
        \label{fig:example_impact_cleaning}
        \label{fig:pipeline_difference_sample_example}
    }
    \caption{High level statistics between two seperate pipelines and a sample example of the difference between two pipelines. First iteration (Figure \ref{fig:pipeline_v0}) generated a 7Mb dataset. After some careful tweaking, and some observed samples, we proposed a new pipeline in order to preserve previously wrongly removed data (Figure \ref{fig:pipeline_v2}) which generated a 134Mb dataset (x18). A example sample is available in Figure \ref{fig:pipeline_difference_sample_example}}
    \label{fig:estimated_sizes}
\end{figure}

As a typical usage of the visualisation tool as a development tool, for documents coming from pseudo-crawls, we wanted to create a method to remove the parts of the documents that looked like a template, based on the principle that these templates would be identifiable by the fact that they would be repeated lines between documents. With the first version of the pipeline we could see from the estimates of the size of the final dataset (Figure \ref{fig:estimated_sizes}) that a lot of content was removed. Looking at the examples (Figure \ref{fig:example_impact_cleaning}), we could confirm that a large part of the article text was removed. The cause of this behaviour was due to the fact that the same article was appearing at several different URLs as the website hierarchy had changed between the different common crawl dumps. For the final pipeline, we therefore added a custom deduplication of the urls as a first operation to target this change of addresses. With the final pipeline developed, less content was removed. By manually inspecting the examples, we could observe that the content removed from the documents was indeed the one initially targeted.

\begin{figure}
    \centering
    % \subfloat{{\includegraphics[width=\textwidth]{figures/appendix/visu_tool_ex_dedup_pseudocrawl\_newspaper_clean_v0.png} }}
    % \quad
    % \subfloat{{\includegraphics[width=\textwidth]{figures/appendix/visu_tool_ex_dedup_pseudocrawl\_newspaper_clean_v2.png} }}
    
\end{figure}

\subsection{Exhaustive list of functions used in (Crowd)Sourced dataset}
\label{appendix:crowdsourced_dataset_exhaustive_list_of_functions}

We provide an exhaustive list of functions used in each of the processing pipeline for the crowdsourced dataset\footnote{Code is available at \href{https://github.com/bigscience-workshop/data-preparation/blob/main/preprocessing/training/clean.py}{https://github.com/bigscience-workshop/data-preparation/blob/main/preprocessing/training/clean.py}}:

% THOMAS: add Angie's google document on simple description of each functions.
% TODO: Thomas fix \n rendering issue
\textbf{replace\_newline\_with\_space} Takes in a batch of texts and for each text replaces the newline character "\\n" with a single space.

\textbf{remove\_lines\_with\_code} Takes in a batch of texts and removes lines with the following substrings: \textit{"\{"}, \textit{"\}"}, \textit{"[if"}, \textit{"<script"},

\textbf{remove\_html\_spans} Takes in a batch of texts and removes lines with the following substrings: 
\textit{"<span"},
\textit{"</span>"},
\textit{"<div"},
\textit{"<a"},
\textit{"</div>"},
\textit{"</a>"},
\textit{"br>"},

\textbf{remove\_html\_spans\_sanad} Takes in a batch of texts and removes lines with the following substrings: 
\textit{"<img"},
\textit{"]]>"},
\textit{"<![CDATA"},
\textit{"//DW"},
\textit{"var "},
\textit{"xtImg"},
\textit{"To view this video please enable JavaScript"},

\textbf{remove\_wiki\_mojibake} Takes in a batch of texts and removes lines with the following substrings: 
\textit{"À À"}

\textbf{strip\_substrings\_en\_wiktionary} Takes in a batch of texts and removes the following substrings:
\begin{itemize}
    \item \textit{This entry needs pronunciation information}
    \item \textit{Please try to find a suitable image on Wikimedia Commons or upload one there yourself!This entry need pronunciation information}
    \item \textit{You may continue to edit this entry while the discussion proceeds, but please mention significant edits at the RFD discussion and ensure that the intention of votes already cast is not left unclear}
    \item \textit{This entry is part of the phrasebook project, which presents criteria for inclusion based on utility, simplicity and commonality}
    \item \textit{If you are a native speaker with a microphone, please record some and upload them}
    \item \textit{If you are familiar with the IPA then please add some!}
    \item \textit{Feel free to edit this entry as normal, but do not remove {{rfv}} until the request has been resolved}
    \item \textit{This entry needs quotations to illustrate usage}
    \item \textit{If you are familiar with the IPA then please add some!This entry needs audio files}
    \item \textit{Please see that page for discussion and justifications}
    \item \textit{If you are familiar with the IPA or enPR then please add some!A user has added this entry to requests for verification(+) If it cannot be verified that this term meets our attestation criteria, it will be deleted}
    \item \textit{This entry needs a photograph or drawing for illustration}
    \item \textit{A user has added this entry to requests for deletion(+)}
    \item \textit{Do not remove the {{rfd}} until the debate has finished}
    \item \textit{This entry needs audio files}
    \item \textit{If you come across any interesting, durably archived quotes then please add them!This entry is part of the phrasebook project, which presents criteria for inclusion based on utility, simplicity and commonality}
    \item \textit{(For audio required quickly, visit WT:APR)}
\end{itemize}

\textbf{remove\_references\_\{lang\}}
Removes lines that do not contain a minimum ratio of stopwords, as defined for each language\footnote{\href{https://github.com/bigscience-workshop/catalogue_data/blob/master/clean_helpers/stopwords.py}{https://github.com/bigscience-workshop/catalogue\_data/blob/master/clean\_helpers/stopwords.py}}. Note, currently does not support languages with different segmentation (e.g. Chinese). Designed for academic datasets.

\textbf{split\_sentences\_\{lang\}}
Builds a sentence splitter depending on the language passed:
For Arabic, Catalan, Basque, Indonesian, and Chinese (both simplified and traditional), we use the Stanza tokenizer \citep{stanza}.
For English, French, Portuguese, and Spanish, we use the NLTK tokenizer \citep{nltk}.
For Bengalic, Gujarati, Hindi, Kannada, Malayalam, Marathi, Punjabi, Tamil, and Telugu, we use the Indic NLP library tokenizer \citep{indic-nlp-library}. 
For Vietnamese, we use the Underthesea tokenizer \footnote{\href{https://github.com/undertheseanlp/underthesea}{https://github.com/undertheseanlp/underthesea}}.

\textbf{filter\_remove\_empty\_docs} 
Removes documents that have a length of 0 when whitespace is removed.

\textbf{filter\_wiki\_user\_titles}
Removes documents where the Wikimedia metadata title starts with \textit{“user”},

\textbf{filter\_wiki\_non\_text\_type}
Removes documents where the Wikimedia metadata type is not “text”

\textbf{filter\_small\_docs}
Discards documents with less than 15 words. Tokenization is done via whitespace tokenizer.

\textbf{filter\_small\_docs\_bytes\_\{i\}}
Discards documents with less than either 300 or 1024 bytes of text

\textbf{dedup\_template\_soft}
Removes lines that are a minimum of 15 characters long and occur 10 or more times.

\textbf{dedup\_pseudocrawl\_newspapers}
Removes lines that occur 2 or more times.

\textbf{dedup\_document}
Removes duplicate documents ignoring whitespaces and punctuation so only keeping characters and keeps one occurrence.

\textbf{dedup\_document\_on\_url}
Removes duplicate documents based on matched url while ignoring query parameters and keeps one occurrence.

\textbf{dedup\_document\_on\_url\_lm\_es\_pseudocrawl-filtered\_341\_es\_cointelegraph\_com}
Removes duplicate documents based on the normalized urls (e.g., \$URL and \$URL/amp are treated as the same) without the query parameters and keeps one occurrence.

\textbf{dedup\_document\_on\_url\_lm\_en\_pseudocrawl\_filtered\_619\_www\_qut\_edu\_au}
Removes duplicate documents based on the url without query parameters except for the "id" and "new-id" query parameters. The "new-id" query parameter is changed into a simple "id" parameter.

\textbf{concatenate\_lm\_fr\_ester}
Concatenate the text sorted by the id number in the metadata.

\section{Exhaustive list of human curated filters used on OSCAR}\label{appendix:oscar_filters}

Before performing the filtering step, we did a cleaning step to modify the documents by standardizing whitespace and removing links, non-printable characters, and long words beyond a character threshold. These steps were designed to remove ``non natural'' language parts of the document (i.e. texts that are machine generated or not language, such as URLs).

Many of these filters require to split a document into words. For Chinese, we used the SentencePiece unigram tokenizer. For Vietnamese, since a word can be composed of two or three sub-words separated by spaces, we augmented the list of space separated tokens by the list of two and three consecutive space separated tokens.

\paragraph{Filter on number of words} We discarded documents with too few words, as they often contain incorrect sentences, or contain no context for a model to learn correctly.

\paragraph{Filter on character repetition ratio} To remove documents containing many repetitions, for a given $n$ (determined in practice according to the language by native speakers), we counted the occurrence of each \textit{character} $n$-gram present in the document. We defined the character repetition ratio as the ratio of the sum of the $k$ largest occurrences by the sum of all occurrences, and we discarded documents with a too high ratio.
\newline
If $k=1$, short sentences are much more likely to have a high character repetition ratio, since the most frequent $n$-gram represents a larger proportion of the sentence. If $k$ is the number of occurrences greater than or equal to $2$, very long documents, but not necessarily including repetitions, tend to have a high character repetition ratio, since these texts inherently have a wide diversity of $n$-grams. We found that $k=\lfloor \sqrt{N} \rfloor $, with $N$ the number of different $n$-grams found in the document, counterbalances well this effect in practice.

\textit{Example:} Take the sentence "\texttt{ok\_ok\_good\_ok}" and $n=3$. Character $n$-grams, with their frequencies, are given in the following table.

\begin{center}
\begin{tabular}{ |c|c|c|c|c|c|c|c|c| } 
 \hline
  \texttt{ok\_} & \texttt{\_ok} & \texttt{k\_o} & \texttt{k\_g} & \texttt{\_go} & \texttt{goo} & \texttt{ood} & \texttt{od\_} & \texttt{d\_o} \\
 \hline
  2 & 2 & 1 & 1 & 1 & 1 & 1 & 1 & 1 \\
  \hline
\end{tabular}
\end{center}

Since we have 9 different character $n$-grams, $N=9$ and $k = \lfloor \sqrt{N} \rfloor =3$.

The sum of the $k$ largest occurrences is $2+2+1=5$ and the sum of all occurrences is $11$. Thus, the character repetition ratio for this sentence is $\frac{5}{11}$.

\paragraph{Filter on word repetition ratio} As a complement to the previous filter, we remove documents that have commonly repeated similar long sentences. More specifically, we create a filter for the repetitions by looking this time at the occurrences of the \textit{word} $n$-grams, for a chosen $n$ parameter. We define the word repetition ratio as the ratio of the sum of the occurrences greater than or equal to 2 to the sum of all occurrences, and we discard documents with too high of a ratio. Contrary to the filter on the character repetition ratios, we did not find a bias of this method giving systematically higher or lower scores to longer or short documents. This filter is more robust in finding documents with long exact duplicated sentences in them, while the previous one is used to find short to medium sized repetitions.

\paragraph{Filter on special character ratio} We established a list of special characters, including emojis, and simply discard documents with a special character ratio above a certain threshold.

\paragraph{Filter on closed class word ratio} We found that having a low closed class word ratio in a document was one of the best indicators of a non-human generated content. We built lists of closed class words for each language by taking pre-existing lists, for example from Universal Dependencies\footnote{\href{https://universaldependencies.org/}{https://universaldependencies.org/}}, which were then reviewed by native speakers. We discard documents with a too low closed class word ratio.

\paragraph{Filter on flagged word ratio} To limit the over-representation of pornographic documents, which are in practice much more likely to have shocking and sexist content, and to contain only buzzwords for SEO, we built lists of flagged words for each language by gathering existing lists, and filtering them by native speakers with precise instructions. We are then able to compute the flagged word ratio of a document and discard it if it is too high. About 1\% of the documents for each language are removed by this filter.

\textit{Instructions for building the lists of flagged words:} Keep only the words associated with porn and systematically used in a sexual context. Remove words that can be used in medical, scientific, colloquial (without referring systematically to porn), or everyday contexts. Remove all insults. Remove all words referring to race or sexual orientation.

\paragraph{Filter on language identification prediction score} We used fastText \citep{joulin2017bag} to perform language identification and getting confidence scores for each document. If a score is below a specific threshold, we discard the document. We chose to eliminate few documents with this filter, because the language identification does not perform as well on low-resource languages.

\paragraph{Filter on perplexity score} Following \citet{wenzek2020ccnet}, we trained SentencePiece unigram tokenizers \citep{kudo-2018-subword} followed by KenLM 5-gram models after tokenization \citep{heafield-2011-kenlm} on Wikipedia article openings for every language that was extracted from OSCAR. As in \citet{BERTIN}, we discarded documents to move the perplexity distribution towards the median, to avoid too high perplexity scores (deemed as not useful for the model), but subsampling was done by perplexity thresholding, not by reshaping the distribution as in \citet{BERTIN}. This thresholding was done lightly, by having native speakers manually establish the cutoff values per language\footnote{Native speakers used an ad-hoc visualization tool built for the occasion: \href{https://huggingface.co/spaces/huggingface/text-data-filtering}{https://huggingface.co/spaces/huggingface/text-data-filtering}}, so as not to be too biased by the Wikipedia content and keep the dataset diverse.

\begin{comment}

\begin{table}[!ht]
\begin{center}
\begin{tabular}{|c|c|c|c|c|c|c|c|c|c|c|c|c|c|} 
 \hline
 & Chinese & English & French & Portuguese & Spanish \\ 
 \hline
 Step 1: Filtering & 10.0 & 10.0 & 10.0 & 10.0 & 10.0 \\ 
 \hline
 Step 2: SimHash deduplication & 10.0 & 10.0 & 10.0 & 10.0 & 10.0 \\ 
 \hline
 Step 3: Substring deduplication & 10.0 & 10.0 & 10.0 & 10.0 & 10.0 \\
 \hline
\end{tabular}
\end{center}
\caption{Percentage (non-cumulative) of discarded documents after the different steps for the five largest languages considered in the common crawl. Step 0: Original OSCAR.}
\end{table}

\begin{table}[!ht]
\small
\begin{center}
\begin{tabular}{|c|c|c|c|} 
 \hline
 & Step 1: Filtering & Step 2: SimHash deduplication & Step 3: Substring deduplication \\
 \hline
 Arabic & 00.0 & 00.0 & 00.0 \\
 \hline
 Basque & 00.0 & 00.0 & 00.0 \\
 \hline
 Bengali & 00.0 & 00.0 & 00.0 \\
 \hline
 Catalan & 00.0 & 00.0 & 00.0 \\
 \hline
 Chinese & 00.0 & 00.0 & 00.0 \\
 \hline
 English & 00.0 & 00.0 & 00.0 \\
 \hline
 French & 00.0 & 00.0 & 00.0 \\
 \hline
 Hindi & 00.0 & 00.0 & 00.0 \\
 \hline
 Indonesian & 00.0 & 00.0 & 00.0 \\
 \hline
 Portuguese & 00.0 & 00.0 & 00.0 \\
 \hline
 Spanish & 00.0 & 00.0 & 00.0 \\
 \hline
 Urdu & 00.0 & 00.0 & 00.0 \\
 \hline
 Vietnamese & 00.0 & 00.0 & 00.0 \\
 \hline
\end{tabular}
\end{center}
\caption{Percentage (non-cumulative) of discarded documents after the different steps for the languages considered in the common crawl. Step 0: Original OSCAR.}
\end{table}

\end{comment}

\section{PII filtering initiative}
\label{sec:muliwai}

Even if not eventually used in our final pipeline, we have released \texttt{muliwai}\footnote{Pronounced \textit{"mu-lee-why"}, Hawaiian for river. \href{https://github.com/ontocord/muliwai/tree/main}{https://github.com/ontocord/muliwai/tree/main}} a library for text pre-processing, augmentation, anonymization, and synthesis. It relies on transformer models and back-translation to perform NER and associated augmentation and anonymization over 100+ languages (i.e., we rely on XLMRoberta \cite{fan2021beyond} and M2M100 \cite{unsupervised-cross-lingual-representation-learning-at-scale}). We either use a specific model for the chosen language or a model with cross-lingual capabilities. \texttt{Muliwai} tags using the aforementioned transformer then translate the sentence to a target language (e.g., English) and test to see if the translation preserves the NER tagging and discounts or increases the weight of a NER decision accordingly. It then performs NER in the target language and back translates to the source language. Finally it matches the translated sentence to the original sentence to determine which text spans in the source language sentence should be NER tagged based on the target language NER. We also use spacy and regex as added signals for NER tags.

We also include in the library specific regexes for detecting age, email, date, time, personal addresses, phone numbers and government-issued identifiers (such as license plates). Some regex matches use also the surrounding text context to improve precision.

However, the scale of the data, the fact that the impact on the resulting text could not be fully assessed in terms of language modeling and the time constraint due to compute allocation, meant this approach could not be operationalized on ROOTS. Instead we fell back to a simpler approach, see Section \ref{sec:pii_regex}.

\section{Data Sources}\label{appendix:datasets}

%\begin{table}
\setlength\tabcolsep{3pt} % default value: 6pt
\begin{center}
\begin{longtable}{|P{0.3\textwidth}|P{0.3\textwidth}|P{0.3\textwidth}|}
\hline
\textbf{Dataset} & \textbf{Language} & \textbf{Source} \\ [0.5ex] 
\hline\hline
\endhead
% \textbf{aggregated sources} & bm, ak, fon, ki, ig, ny, rw, sn, st, rn, nso, ln, lg, tn, zu, tum, ts, wo, tw, sw, yo, xh &  \\
\hline
\textbf{AraBench} & ar & \cite{arabench} \\
\hline
\textbf{1.5 billion words Arabic Corpus} & ar & \cite{arabic-billion-words} \\
\hline
\textbf{BanglaLM} & bn &  \cite{bangla-lm} \\
\hline
\textbf{bangla sentiment classification datasets} & bn & \cite{bangla-sentiment-classification-datasets} \\
\hline
\textbf{Question answering in Bengali} & bn & \cite{bengali-question-answering} \\
\hline
\textbf{Binhvq News Corpus} & vi & \cite{binhvq-news-corpus} \\
\hline
\textbf{Books by Book Dash} & en, fr, xh, zu & \href{https://bookdash.org/books/}{https://bookdash.org/books/} \\
\hline
\textbf{Bloom Library} & ak, bm, fon, ki, lg, ln, nso, rw, st, sw, tn, ts, xh, zu  & \href{https://bloomlibrary.org/}{bloomlibrary.org} \\
\hline
\textbf{BRAD 2.0} & ar & \cite{brad-2} \\
\hline
\textbf{brWaC Corpus} & pt & \\
\hline
\textbf{EusCrawl} & eu & \cite{bsbasque} \\
\hline
\textbf{Catalan General Crawling} & ca & \cite{catalan-crawling-xquad-ca} \\
\hline
\textbf{Catalan Government Crawling} & ca & \cite{catalan-crawling-xquad-ca} \\
\hline
\textbf{Catalan Textual Corpus} & ca & \cite{catalan-crawling-xquad-ca}  \\
\hline
\textbf{Data on COVID-19 News Coverage in Vietnam} & vi & \cite{data-on-covid-19-news-coverage-in-vietnam} \\
\hline
\textbf{DuReader} & zhs & \cite{du-reader} \\
\hline
\textbf{Enriched CONLLU Ancora for ML training} & ca & \cite{enriched-conllu-ancora-for-ml-training} \\
\hline
\textbf{ESTER} & fr & \cite{ester} \\
\hline
\textbf{Github} & code & \href{https://cloud.google.com/blog/topics/public-datasets/github-on-bigquery-analyze-all-the-open-source-code}{Github on BigQuery} \\
\hline
\textbf{Habibi} & ar & \cite{habibi} \\
\hline
\textbf{HAL} & fr &  \href{https://hal.archives-ouvertes.fr/}{hal.archives-ouvertes.fr/} \\
\hline
\textbf{IIT Bombay English-Hindi Parallel Corpus} & hi & \cite{iitb-english-hindi-corpus} \\
\hline
\textbf{IndicNLP Corpus} & bn, gu, hi, kn, ml, mr, or, pa, ta, te & \cite{indic-nlp-corpus} \\
\hline
\textbf{Indo4B BPPT} & id & \cite{indo4b-bppt} \\
\hline
\textbf{Indo4B OPUS JW300} & id & \cite{indo4b-jw300} \\
\hline
\textbf{Indo4B Kompas} & id & \cite{indo4b-tempo-kompas} \\
\hline
\textbf{Indo4B Parallel Corpus} & id & \cite{indo4b-parallel} \\
\hline
\textbf{Indo4B TALPCo} & id & \cite{indo4b-talpco} \\
\hline
\textbf{Indo4B Tempo} & id & \cite{indo4b-tempo-kompas} \\
\hline
\textbf{Indonesian Frog Storytelling corpus} & id & \cite{indonesian-frog-storytelling-corpus} \\
\hline
\textbf{Indonesian News Articles Published at 2017} & id & \cite{indonesian-news-articles-2017} \\
\hline
\textbf{Indonesian News Corpus} & id & \cite{indonesian-news-corpus} \\
\hline
\textbf{IndoNLI} & id & \cite{indonli} \\
\hline
\textbf{Indosum} & id & \cite{indosum} \\
\hline
\textbf{KALIMAT} & ar & \cite{kalimat} \\
\hline
\textbf{KSUCCA} & ar & \cite{ksucca} \\
\hline
\textbf{LABR} & ar & \cite{labr} \\
\hline
\textbf{Language modeling data for Swahili} & sw & \cite{hf-swahili} \\
\hline
\textbf{Leipzig Corpora Collection} & ur & \cite{leipzig-wortschatz} \\
\hline
\textbf{Mann Ki Baat} & bn, gu, hi, ml, mr, or, ta, te, ur & \cite{mkb-pib} \\
\hline
\textbf{Masakhaner} & ig, lg, rw, sw, wo, yo & \cite{masakhaner} \\
\hline
\textbf{MultiUN v2} & en, ar, es, fr, zhs & \cite{multi-un-2} \\
\hline
% Thomas: Removing as this is the same as Stack Exchange
% \textbf{no code stackexchange} & en & \cite{Gao2020} \\
% \hline
\textbf{Odiencorp} & en, or & \cite{odiencorp} \\
\hline
\textbf{Opensubtitles2016} & ca, en, ar, es, eu, fr, id, bn, hi, ml, ta, te, ur, pt, vi, zhs & \cite{open-subtitles} \\
\hline
\textbf{OpenITI proc} & ar & \cite{openiti-proc} \\
\hline
\textbf{OPUS-100} & ca, ar, eu, id, as, bn, gu, hi, ig, kn, ml, mr, or, pa, rw, ta, te, ur, pt, vi, xh, yo, zu & \cite{opus100} \\
\hline
\textbf{ParlamentParla} & ca & \cite{parlament-parla} \\
\hline
\textbf{PIB} & bn, gu, hi, ml, mr, or, pa, ta, te, ur & \cite{mkb-pib} \\
\hline
\textbf{Project Gutenberg} & en, es, fr, pt, zhs & \href{https://www.gutenberg.org/}{gutenberg.org} \\
\hline
\textbf{QED (formely AMARA Corpus)} & ar, en, es, fr, hi, pt, zhs, zht & \cite{qedcorpus} \\
\hline
\textbf{Recibrew} & id & \cite{recibrew} \\
\hline
\textbf{The Royal Society Corpus} & en & \cite{royal-society-corpus} \\
\hline
\textbf{S2ORC} & en & \cite{lo-wang-2020-s2orc} \\
\hline
\textbf{Samanantar} & as, bn, gu, hi, kn, ml, mr, or, pa, ta, te & \cite{samanantar} \\
\hline
\textbf{SANAD} & ar & \cite{sanad} \\
\hline
\textbf{SciELO} & en, es, pt &  \href{https://scielo.org/}{scielo.org}\\
\hline
\textbf{Stack Exchange} & code & \cite{Gao2020} \\
\hline
\textbf{Swahili News Classification Dataset} & sw & \cite{swahili-news} \\
\hline
\textbf{Tashkeela} & ar & \cite{tashkeela} \\
\hline
\textbf{TeCLa} & ca & \cite{tecla} \\
\hline
\textbf{WIT$^3$} & ca, ar, en, es, eu, fr, id, as, bn, gu, hi, kn, ml, mr, pa, sw, ta, te, ur, pt, vi, zhs & \cite{ted-talks-iwslt} \\
\hline
\textbf{The Pile: EuroParl} & en, es, fr, pt & \cite{Gao2020} \\
\hline
\textbf{The Pile: USPTO} & en & \cite{Gao2020} \\
\hline
\textbf{UIT-VSMEC} & vi & \cite{uit-vsmec} \\
\hline
\textbf{United Nations Parallel Corpus} & ar, en, es, fr, zhs & \cite{uncorpus} \\
\hline
\textbf{Unsupervised Cross-lingual Representation Learning at Scale Common Crawl Corpus} & ne &  \cite{unsupervised-cross-lingual-representation-learning-at-scale} \\
\hline
\textbf{Urdu Monolingual Corpus} & ur & \cite{urdu-monolingual-corpus} \\
\hline
\textbf{VietAI SAT} & vi & \cite{vietai-sat} \\
\hline
\textbf{Vietnamese Poetry Corpus} & vi & \cite{vietnamese-poetry} \\
\hline
\textbf{UIT-VSFC} & vi & \cite{vietnamese-students-feedback} \\
\hline
\textbf{VilaQuAD} & ca & \cite{vilaquad} \\
\hline
\textbf{VinBigdata-VSLP ASR Challenge 2020} & vi & \href{https://institute.vinbigdata.org/events/vinbigdata-chia-se-100-gio-du-lieu-tieng-noi-cho-cong-dong/}{institute.vinbigdata.org/events/vinbigdata-chia-se-100-gio-du-lieu-tieng-noi-cho-cong-dong/}\\
\hline
\textbf{VinBigdata-VSLP Monolingual Corpus 2020} & vi & \href{https://institute.vinbigdata.org/events/vinbigdata-chia-se-100-gio-du-lieu-tieng-noi-cho-cong-dong/}{institute.vinbigdata.org/events/vinbigdata-chia-se-100-gio-du-lieu-tieng-noi-cho-cong-dong/} \\
\hline
\textbf{VinBigdata-VSLP Bilingual Corpus 2020} & vi & \href{https://institute.vinbigdata.org/events/vinbigdata-chia-se-100-gio-du-lieu-tieng-noi-cho-cong-dong/}{institute.vinbigdata.org/events/vinbigdata-chia-se-100-gio-du-lieu-tieng-noi-cho-cong-dong/} \\
\hline
\textbf{ViquiQuAD} & ca & \cite{viquiquad} \\
\hline
\textbf{VNTQcorpus(big)} & vi & \href{viet.jnlp.org/download-du-lieu-tu-vung-corpus}{http://viet.jnlp.org/download-du-lieu-tu-vung-corpus}\\
\hline
\textbf{Wikibooks} & ca, ar, en, es, eu, fr, id, bn, hi, ml, mr, pa, ta, te, ur, pt, vi, zhs & \href{https://www.wikibooks.org/}{wikibooks.org} \\
\hline
\textbf{Wikimedia} & ca, id, hi, pt & \href{https://www.wikimedia.org/}{wikimedia.org} \\
\hline
\textbf{Wikinews} & ar, ca, en, es, fr, ta, pt, zhs & \href{https://www.wikinews.org/}{wikinews.org} \\
\hline
\textbf{Wikipedia} & ak, ar, as, bm, bn, ca, en, es, eu, fr, id, ig, gu, hi, ki, kn, lg, ln, ml, mr, nso, ny, or, pa, pt, rn, rw, sn, st, sw, ta, te, tn, ts, tum, tw, ur, vi, wo, yo, zhs, zht, zu & \href{https://www.wikipedia.org/}{wikipedia.org} \\
\hline
\textbf{Wikiquote} & ar, ca, en, es, eu, fr, id, gu, hi, kn, ml, mr, ta, te, ur, pt, vi, zhs & \href{https://www.wikiquote.org/}{wikiquote.org} \\
\hline
\textbf{Wikisource} & ar, ca, es, eu, fr, id, as, bn, gu, hi, kn, ml, mr, or, pa, ta, te, pt, vi & \href{https://wikisource.org/}{wikisource.org} \\
\hline
\textbf{Wikiversity} & ar, en, es, fr, hi, pt, zhs & \href{https://www.wikiversity.org/}{wikiversity.org} \\
\hline
\textbf{Wikivoyage} & en, es, fr, bn, hi, pt, vi, zhs & \href{https://www.wikivoyage.org/}{wikivoyage.org} \\
\hline
\textbf{Wiktionary} & ar, ca, en, es, eu, fr, id, as, bn, gu, hi, kn, ml, mr, or, pa, ta, te, ur, pt, vi & \href{https://www.wiktionary.org/}{wiktionary.org} \\
\hline
\textbf{WuDaoCorpora} & zhs & \cite{yuan2021wudaocorpora} \\
\hline
\textbf{XQUAD-ca} & ca &  \cite{catalan-crawling-xquad-ca} \\
\hline
\caption{List of datasets used in crowdsourced dataset.}
\label{tab:list_crowd_source_dataset}
\end{longtable}
\end{center}
%\end{table}

\newpage
\begin{table}[]
\centering
\resizebox{\textwidth}{!}{%
\begin{tabular}{lTTlllr}
\toprule
\textbf{Language}            & \textbf{ISO-639-3} & \textbf{catalog-ref} & \textbf{Genus}                   & 
\textbf{Family}         & \textbf{Macroarea} & \textbf{Size in Bytes} \\
\midrule
Akan                & aka       & ak          & Kwa                     & Niger-Congo    & Africa    & 70,1554        \\
Arabic              & arb       & ar          & Semitic                 & Afro-Asiatic   & Eurasia   & 74,854,900,600   \\
Assamese            & asm       & as          & Indic                   & Indo-European  & Eurasia   & 291,522,098     \\
Bambara             & bam       & bm          & Western Mande           & Mande          & Africa    & 391,747        \\
Basque              & eus       & eu          & Basque                  & Basque         & Eurasia   & 2,360,470,848    \\
Bengali             & ben       & bn          & Indic                   & Indo-European  & Eurasia   & 18,606,823,104   \\
Catalan             & cat       & ca          & Romance                 & Indo-European  & Eurasia   & 17,792,493,289   \\
Chi Chewa           & nya       & ny          & Bantoid                 & Niger-Congo    & Africa    & 1,187,405       \\
Chi Shona           & sna       & sn          & Bantoid                 & Niger-Congo    & Africa    & 6,638,639       \\
Chi Tumbuka         & tum       & tum         & Bantoid                 & Niger-Congo    & Africa    & 170,360        \\
English             & eng       & en          & Germanic                & Indo-European  & Eurasia   & 484,953,009,124  \\
Fon                 & fon       & fon         & Kwa                     & Niger-Congo    & Africa    & 2,478,546       \\
French              & fra       & fr          & Romance                 & Indo-European  & Eurasia   & 208,242,620,434  \\
Gujarati            & guj       & gu          & Indic                   & Indo-European  & Eurasia   & 1,199,986,460    \\
Hindi               & hin       & hi          & Indic                   & Indo-European  & Eurasia   & 24,622,119,985   \\
Igbo                & ibo       & ig          & Igboid                  & Niger-Congo    & Africa    & 14078,521      \\
Indonesian          & ind       & id          & Malayo-Sumbawan         & Austronesian   & Papunesia & 19,972,325,222   \\
Isi Zulu            & zul       & zu          & Bantoid                 & Niger-Congo    & Africa    & 8,511,561       \\
Kannada             & kan       & kn          & Southern Dravidian      & Dravidian      & Eurasia   & 2,098,453,560    \\
Kikuyu              & kik       & ki          & Bantoid                 & Niger-Congo    & Africa    & 359,615        \\
Kinyarwanda         & kin       & rw          & Bantoid                 & Niger-Congo    & Africa    & 40,428,299      \\
Kirundi             & run       & rn          & Bantoid                 & Niger-Congo    & Africa    & 3,272,550       \\
Lingala             & lin       & ln          & Bantoid                 & Niger-Congo    & Africa    & 1,650,804       \\
Luganda             & lug       & lg          & Bantoid                 & Niger-Congo    & Africa    & 4,568,367       \\
Malayalam           & mal       & ml          & Southern Dravidian      & Dravidian      & Eurasia   & 3,662,571,498    \\
Marathi             & mar       & mr          & Indic                   & Indo-European  & Eurasia   & 1,775,483,122    \\
Nepali              & nep       & ne          & Indic                   & Indo-European  & Eurasia   & 2,551,307,393    \\
Northern Sotho      & nso       & nso         & Bantoid                 & Niger-Congo    & Africa    & 1,764,506       \\
Odia                & ori       & or          & Indic                   & Indo-European  & Eurasia   & 1,157,100,133    \\
Portuguese          & por       & pt          & Romance                 & Indo-European  & Eurasia   & 79,277,543,375   \\
Punjabi             & pan       & pa          & Indic                   & Indo-European  & Eurasia   & 1,572,109,752    \\
Sesotho             & sot       & st          & Bantoid                 & Niger-Congo    & Africa    & 751,034        \\
Setswana            & tsn       & tn          & Bantoid                 & Niger-Congo    & Africa    & 1,502,200       \\
Simplified Chinese  &     ---      & zhs         & Chinese                 & Sino-Tibetan   & Eurasia   & 261,019,433,892  \\
Spanish             & spa       & es          & Romance                 & Indo-European  & Eurasia   & 175,098,365,045  \\
Swahili             & swh       & sw          & Bantoid                 & Niger-Congo    & Africa    & 236,482,543     \\
Tamil               & tam       & ta          & Southern Dravidian      & Dravidian      & Eurasia   & 7,989,206,220    \\
Telugu              & tel       & te          & South-Central Dravidian & Dravidian      & Eurasia   & 2993407,159    \\
Traditional Chinese &      ---     & zht         & Chinese                 & Sino-Tibetan   & Eurasia   & 762,489,150     \\
Twi                 & twi       & tw          & Kwa                     & Niger-Congo    & Africa    & 1,265,041       \\
Urdu                & urd       & ur          & Indic                   & Indo-European  & Eurasia   & 2,781,329,959    \\
Vietnamese          & vie       & vi          & Viet-Muong              & Austro-Asiatic & Eurasia   & 43,709,279,959   \\
Wolof               & wol       & wo          & Wolof                   & Niger-Congo    & Africa    & 3,606,973       \\
Xhosa               & xho       & xh          & Bantoid                 & Niger-Congo    & Africa    & 14,304,074      \\
Xitsonga            & tso       & ts          & Bantoid                 & Niger-Congo    & Africa    & 707,634        \\
Yoruba              & yor       & yo          & Defoid                  & Niger-Congo    & Africa    & 89,695,835      \\
Programming Languages                & ---      & ---        & ---                    & ---           &       & 174,700,245,772 \\
\bottomrule \\
\end{tabular}%
}
\caption{Linguistic makeup of the corpus.}
\label{tab:language_families}
\end{table}

\begin{table}[!htb]
\centering
\begin{tabular}{lcr}
\toprule
\textbf{Language}     & \textbf{Number of Pseudocrawled Domains} & \textbf{Size in Bytes}       \\
\midrule
Spanish       & 108                             & 29,440,210,712 \\
English       & 22                              & 4,537,031,408  \\
Swahili & 5 & 109,110,002 \\
Indonesian      & 4                               & 770,023,233   \\
Basque       & 4                               & 281,610,312   \\
French       & 3                               & 1,416,682,404  \\
Hindi & 2                               & 1,536,649,276  \\
Simplified Chinese      & 2                               & 173,884,238   \\
Yoruba & 2 & 6,198,347 \\
Igbo & 2 & 2,650,116 \\
Arabic       & 1                               & 694,455,304   \\
Portuguese       & 1                               & 30,615,557   \\
Kinyarwanda & 1 & 9,301,301 \\
\bottomrule \\
\end{tabular}%

\caption{Pseudocrawled data per language sorted by number of domains crawled}
\label{tab:pseudocrawled_data}
\end{table}

\clearpage
\newpage

\section{Author contributions}

Author contributions in alphabetical order.

\textbf{Aaron Gokaslan} set a \texttt{pre-commit} (for code formatting) in a repository and helped with the writing of the Related Work section of the paper.

\textbf{Aitor Soroa} integrated one dataset into crowdsourced data.

\textbf{Albert Villanova del Moral} led the gathering of identified sources, implemented loading scripts in the \textit{datasets} library in a single unified interface, and integrated the most datasets into crowdsourced data.

\textbf{Angelina McMillan-Major} gathered the lists of closed class words for many languages used for the filtering of OSCAR.

\textbf{Anna Rogers} contributed to the writing of the paper.

\textbf{Chenghao Mou} was the main contributor for OSCAR deduplication.

\textbf{Christopher Akiki} advised on analysis aspects of the project, integrated over a hundred datasets into crowdsourced data, wrote dataset loading scripts, participated in cleaning and filtering efforts, helped with visualization, and contributed to the writing of the paper.

\textbf{Daniel van Strien} integrated one dataset into crowdsourced data.

\textbf{David Ifeoluwa Adelani} participated in the PII filtering initiative (see Appendix \ref{sec:muliwai}).

\textbf{Eduardo González Ponferrada} trained SentencePiece and KenLM models used for the filtering of OSCAR.

\textbf{Francesco De Toni} led one crowdsourcing hackathon, analyzed the distribution of the sources in the catalogue, participated in the PII filtering initiative (see Appendix \ref{sec:muliwai}), and contributed to the writing of the paper.

\textbf{Giada Pistilli} helped write the Ethical Considerations section of the paper.

\textbf{Gérard M Dupont} contributed to data tooling and sourcing and advised on analysis aspects of the project.

\textbf{Hieu Tran} helped set the filtering parameters for Vietnamese and contributed to the list of Vietnamese closed-class words.

\textbf{Hugo Laurençon} developed the filtering library used for the cleaning of OSCAR and the visualization tool to help choose the filtering parameters, and ran OSCAR filtering jobs. He was involved in the cleaning of some crowdsourced and pseudo-crawled datasets and the deduplication of OSCAR. He also contributed to the writing of the paper.

\textbf{Huu Nguyen} contributed to the data tooling and lead the PII filtering initiative (see Appendix \ref{sec:muliwai}).

\textbf{Ian Yu}  participated in the PII filtering initiative  (see Appendix \ref{sec:muliwai}) and helped to choose the filtering parameters for Chinese.

\textbf{Itziar Gonzalez-Dios}, as a Basque native speaker, helped choose the filtering parameters for this language.

\textbf{Javier De la Rosa} contributed with perplexity sampling efforts for OSCAR (not used in final pipeline).

\textbf{Jenny Chim} participated in the PII filtering initiative  (see Appendix \ref{sec:muliwai}) and helped to choose the filtering parameters and closed-class words for Chinese.

\textbf{Jian Zhu} integrated two datasets into crowdsourced data.

\textbf{Jörg Frohberg} integrated multiple datasets into crowdsourced data and reached out to license holders.

\textbf{Khalid Almubarak} integrated some datasets into crowdsourced data.

\textbf{Kyle Lo} integrated one dataset into crowdsourced data.

\textbf{Leandro von Werra} participated in cleaning and filtering efforts, built the code dataset, contributed to its analysis, and participated in the sourcing effort for target datasets.

\textbf{Leon Weber} integrated one dataset into crowdsourced data.

\textbf{Long Phan} participated in the PII filtering initiative (see Appendix \ref{sec:muliwai}).

\textbf{Loubna Ben allal} contributed the analysis of the code dataset.

\textbf{Lucile Saulnier} co-led pseudo-crawled data acquisition, contributed to filtering, cleaning, and deduplication for the crowdsourced datasets, built visualization tools to inspect the results of pre-processing, scaled the PII filtering process, performed the document size analysis, and participated in paper writing.

\textbf{Manan Dey}, as a Bengali native speaker, helped to choose the filtering parameters for this language.

\textbf{Manuel Romero Mu\~{n}oz} contributed to KenLM models and to corpus visualization with those models, and participated in the PII filtering initiative (see Appendix \ref{sec:muliwai}).

\textbf{Maraim Masoud} contributed to the sourcing of some Arabic datasets, the list of Arabic closed-class words, and the writing of the paper.

\textbf{Margaret Mitchell} co-led the final regex-based PII efforts.

\textbf{Mario Šaško} integrated multiple datasets into crowdsourced data.

\textbf{Olivier Nguyen} helped build the first blocks of the OSCAR filtering pipeline.

\textbf{Paulo Villegas} participated in the PII filtering initiative (see Appendix \ref{sec:muliwai}) and the perplexity sampling efforts for OSCAR.

\textbf{Pedro Ortiz Suarez} contributed to crowdsourced data and provided high-level metrics on OSCAR.

\textbf{Pierre Colombo} participated in the PII filtering initiative (see Appendix \ref{sec:muliwai}).

\textbf{Quentin Lhoest} integrated multiple datasets into crowdsourced data.

\textbf{Sasha Luccioni} co-led the final regex-based PII efforts. participated in filtering and cleaning efforts, and contributed to the writing of the paper.

\textbf{Sebastian Nagel} helped to implement the pseudo-crawled data acquisition step.

\textbf{Shamik Bose} participated in the PII filtering initiative (see Appendix \ref{sec:muliwai}) and contributed the list of Bengali closed-class words.

\textbf{Shayne Longpre} contributed to the writing of the paper.

\textbf{Somaieh Nikpoor} co-led the development of ethical/legal charter and contributed to the section on ethical considerations.

\textbf{Stella Biderman} contributed to the writing of the paper.

\textbf{Suhas Pai} participated in the PII filtering initiative (see Appendix \ref{sec:muliwai}).

\textbf{Suzana Ilić} coordinated the organization of BigScience working groups.

\textbf{Teven Le Scao} led final quality control checks, contributed to filtering, cleaning, and deduplication for all components of the corpus, contributed to the OSCAR filtering visualization tool, contributed and repaired several datasets for crowdsourced data, performed the tokenizer-based analysis, and participated in the writing of the paper.

\textbf{Thomas Wang} co-led pseudo-crawled data acquisition, built the distributed cleaning pipelines for pseudo-crawled and crowdsourced datasets, handled job monitoring for crowdsourced dataset filtering, and participated in paper writing.

\textbf{Tristan Thrush} participated in the final regex-based PII efforts.

\textbf{Violette Lepercq} was the primary project manager for the final dataset cleaning efforts and helped to reach out to the native speakers to tune the filtering parameters.

\textbf{Vu Minh Chien} participated in the PII filtering initiative (see Appendix \ref{sec:muliwai}).

\textbf{Yacine Jernite} defined the primary goals of the project, advised on data collection and filtering efforts, contributed sourcing tools and early cleaning scripts, and contributed to the writing of the paper.

\textbf{Zaid Alyafeai} helped find a list of Arabic datasets that should be integrated into crowdsourced data.

% Todo: 
% - Pile's paper figure about tokens: pile tokenization figure
% - distribution of langues
% - table of top sources (% in final dataset) -> even pie chart
% - 2 other figures
% - creating a series of spaces (should be done for the supplementary material) :
%     - clustering space
%     - tokenizer top words

\end{document}